\documentclass{discpaper}
\usepackage[ansinew]{inputenc}
\usepackage{amsmath,amssymb}
\usepackage[ngerman,english]{babel}
\usepackage{multicol}
\usepackage{multirow}
\usepackage{rotating}

\def\IR{{\rm I\!R}}

\newtheorem{proposition}{Proposition}[section]
\newtheorem{corollary}{Corollary}[section]
\newtheorem{theorem}{Theorem}[section]
\newtheorem{acknowledgements}{Acknowledgements}

\begin{document}
\discno{1/12}
\title{Fast nonparametric classification based on data depth}
\author{Tatjana Lange\thanks{Fachbereich Informatik und Kommunikationssysteme,
Hochschule Merseburg, D-06217 Merseburg}
\and Karl Mosler\thanks{Seminar für Wirtschafts-- und Sozialstatistik,
Universität zu Köln, D-50923 Köln}
\and Pavlo Mozharovskyi\thanks{Seminar für Wirtschafts-- und Sozialstatistik,
Universität zu Köln, D-50923 Köln}
\and
\and
\and Revised December 2012
}
\maketitle

\begin{abstract}
A new procedure, called $DD\alpha$-procedure, is developed to solve the problem of classifying $d$-dimensional objects into $q\ge 2$ classes. The procedure is completely nonparametric; it uses $q$-dimensional depth plots and a very efficient algorithm for discrimination analysis in the depth space $[0,1]^q$. Specifically, the depth is the zonoid depth, and the algorithm is the $\alpha$-procedure. In case of more than two classes several binary classifications are performed and a majority rule is applied.
Special treatments are discussed for `outsiders', that is, data having zero depth vector.
The DD$\alpha$-classifier is applied to simulated as well as real data, and the results are compared with those of similar procedures that have been recently proposed.
In most cases the new procedure has comparable error rates, but is much faster than other classification approaches, including the SVM.
\end{abstract}


\newpage



\section{Introduction}\label{intro}

A steady interest in statistical learning theory has intensified recently since nonparametric tools have become available. A new impetus has been given to supervised classification by employing depth functions such as Tukey's (\cite{Tukey75}) halfspace depth or Liu's (\cite{Liu90}) simplicial depth.
In supervised learning a function is constructed from labeled training data that classifies an arbitrary data point by assigning it one of the labels \cite{HastieTF09}.
Given two or more labeled clouds of training data in $d$-space, a data depth measures the centrality of a point with respect to these clouds. For any point in $d$-space it indicates the degree of closeness to each label. This can be employed in different ways for solving the classification task.
Many authors have made use of data depth ideas in supervised classification.
Liu et al. \cite{LiuPS99} were the first who stressed the usefulness and versatility of depth transformations in multivariate analysis. They introduced the notion of a DD-plot, that is the two-dimensional representation of multivariate objects by their data depths regarding two given distributions.
In a straightforward way, an object can be classified to the class where it is deepest, that is, according to its maximum depth. Jornsten \cite{Jornsten04} and Ghosh and Chaudhuri \cite{GhoshC05b} have followed this and similar approaches; see also Hoberg and Mosler \cite{HobergM06}.
Dutta and Ghosh \cite{DuttaG11a,DuttaG11b} 
employ a separator that is linear in a density based on kernel estimates of the projection depth, respectively $L_p$-depth.
Recently, Li et al. \cite{LiCAL11}
have used polynomial separators of the DD-plot to classify objects by their depth representation. These methods differ in the notion of depth used and allow for adaptive and other extensions.

The quoted literature has in common that a (possibly high-dimensional) space of objects is transformed into a lower-dimensional space of depth values of these objects and the classification task is performed in the depth space.
In this context several questions arise:
\begin{enumerate}
\item Which particular notion of depth should be employed?
\item Which classification procedure should be applied to the depth-represented data?
\item How extends the procedure to $q > 2$ classes?
\end{enumerate}

The above literature answers these questions in different ways.
Ad (1), halfspace and simplicial depths, among others, have been employed in \cite{GhoshC05a,LiCAL11,LiuPS99}. They depend only on the combinatorial structure of the data, being constant in the compartments spanned by them. Consequently, these depths are rather robust to outlying data, but calculating them
in higher dimensions can be cumbersome if not impossible. On the other hand Mahalanobis depth  \cite{Mahalanobis36}, which has also been used by these authors, is easily calculated but highly non-robust. Moreover, it depends on the first two moments only and does not reflect any asymmetries of the data. More robust forms of the Mahalanobis depth remain still insensitive to data asymmetries. $L_1$-depth as used in \cite{Jornsten04} has similar drawbacks. \cite{DuttaG11b} employ $L_p$-depths, which are easily calculated if $p$ is known, and choose $p$ in an adaptive procedure; however the latter needs heavy computations. In \cite{HobergM06} the maximum zonoid depth and a combination of it with the Mahalanobis depth are used; both can be efficiently calculated also in high dimensions but lack robustness.
Ad (2), Li et al. \cite{LiCAL11} solve the classification problem of the DD-plot by designing a polynomial line that separates the unit square and provides a minimal average misclassification rate (AMR); the order (up to three) of the polynomial is selected by cross validation.
Similarly, separators are determined in \cite{DuttaG11a} and \cite{DuttaG11b} by cross-validation.

Ad (3) with $q>2$ classes a given point is usually classified in two steps according to majority rule: firstly $\binom{q}{2}$ classifications are performed that are restricted to pairs of classes in the object space, and secondly the point is assigned to that class where it was most often assigned in step 1.

In this paper, ad (1), we employ the zonoid depth \cite{KoshevoyM97,Mosler02}, as it can be efficiently calculated also in higher dimensions (up to $d=20$ and more) and has excellent theoretical properties regarding continuity and statistical inference. However the zonoid depth has a low breakdown point.  If, in a concrete application, robustness is an issue the data have to be preprocessed by some outlier detection procedure.
Ad (2), for final classification in the depth space a variant of the $\alpha$-procedure is employed. It operates simply and very efficiently on low-dimensional spaces like the depth spaces considered here. The $\alpha$-procedure has been originally developed by Vasil'ev \cite{Vasilev91,Vasilev03} and Lange \cite{VasilevL98}.
Ad (3) we employ DD-plots if there are two classes and $q$-dimensional depth plots if there are $q>2$ classes. Assignment of a given point  to a class is based on $\binom{q}{2}$ binary classifications in the $q$-dimensional depth space plus a majority rule. Note that in each binary classification the whole depth information regarding all $q$ classes is used.

We call our approach the DD$\alpha$-approach and apply it to simulated as well as real data. The results are contrasted with those obtained in
\cite{LiCAL11}, \cite{DuttaG11a}, and \cite{DuttaG11b}.

The contribution of this paper is threefold. A classification procedure is proposed that
\begin{enumerate}
  \item is efficiently computable for objects of higher dimensions,
  \item employs a very fast classification procedure of the D-transformed data,
  \item uses the full multivariate information when classifying into $q>2$ classes,
\end{enumerate}

The rest of the paper is organized as follows. Section 2 introduces the depth transform, which maps the data from $d$-dimensional object space to $q$-dimensional depth space, and provides a first discussion of the problem of `outsiders', that are points having a vanishing depth vector. In Section 3 our modification of the $\alpha$-procedure is presented in some detail. Section 4 provides a number of theoretical results regarding the behavior of the DD$\alpha$-procedure on elliptical and mirror symmetric distributions. Section 5 contains extensive simulation results and comparisons. Calculations of real data benchmark examples are reported in Section 6 as well as a comparison of the DD$\alpha$-procedure with the SVM approach. Section 7 concludes.

\section{Depth transform}\label{ddplot}

A data depth is a function that measures, in a certain sense, how close a given point ${\bf x}$ is located to the ``center'' of a finite set $X$ in $\mathbb{R}^d$, that is, how ``deep'' it is in the set. More precisely, a data depth is a function
\[({\bf x}, X) \mapsto D_X({\bf x}) \in [0,1]\,, \quad {\bf x}\in \mathbb{R}^d\,, \quad X\subset \mathbb{R}^d\,,
\]
that satisfies the following restrictions: affine invariant;
upper semicontinuous in $\bf x$; quasiconcave in $\bf x$ (that is, having convex upper level sets), vanishing if $||{\bf x}|| \to \infty$.
 Sometimes two weaker restrictions are imposed: orthogonal invariant; decreasing on rays from a point of maximal depth (that is, starshapedness of the upper level sets).
 For surveys of these restrictions and many special notions of data depth, see e.g. \cite{ZuoS00,Mosler02,Dyckerhoff04,Serfling06,Cascos09}.

Now, assume that data in $\mathbb{R}^d$ are to be classified into $q\ge 2$ classes and that $X_1,\dots, X_q\subset \mathbb{R}^d$ are training sets for these classes each having finite size $n_j=|X_j|$. Let $D$ be a data depth. The function $\mathbb{R}^d\to [0,1]^q$ mapping
\begin{equation}\label{depthtransform}
  {\bf x} \mapsto {\bf d} := (D_{X_1}({\bf x}),\dots, D_{X_q}({\bf x}))
\end{equation}
will be mentioned as a \textit{depth representation}. Each object is represented by a vector whose $q$ components indicate its depth or closeness regarding the $q$ classes. In particular, the training sets $X_j\subset \mathbb{R}^d$ are transformed to sets in  $[0,1]^q$ that represent the classes in the depth space. It should be noted that `closeness' of points in the original space translates to `closeness' of their representations. The classification problem then becomes one of partitioning the depth space $[0,1]^q$ into $q$ parts.

A simple rule, e.g., is to classify a point to that class where it has the largest depth value; see \cite{GhoshC05b,Jornsten04}. This means that the depth space decomposes into $q$ compartments which are separated by (parts of) $q$ bisecting hyperplanes. Maximum depth classification is a linear rule.
A nonlinear classification rule is used in Li et al. \cite{LiCAL11},
who treat the case $q=2$ by constructing a polynomial line up to degree 3 that separates the depth space $[0,1]^2$; see also
\cite{DuttaG11a,DuttaG11b}.

With several important notions of data depth,  $D_X(x)$ vanishes outside the convex hull of $X$. This is, e.g., the case with the halfspace, simplicial, and zonoid depths, but not with the Mahalanobis and $L_p$-depths.
A point that is not within the convex hull of at least one training set then is mapped to the origin in the depth space. Such a point will be mentioned as an \textit{outsider}. Of course, it can be neither regarded as correctly classified nor ignored.
To classify this point we may consider three principal approaches, each allowing for several variants.
  \begin{itemize}

  \item Classify randomly, with probabilities equal to the expected proportions of origin of points to be classified.
    \item Use the $k$-nearest neighbors method with a properly chosen distance: Euclidean distance, $L_p$-distance, Mahalanobis distance with moment estimates, Mahalanobis distance with robust estimates (MCD, cf.\ e.g.\ \cite{HubertD04}).
    \item Classify with maximum Mahalanobis depth (using moment estimates or MCD) or with the maximum of another depth that is properly extended beyond the convex hull as e.g.\ in \cite{HobergM06}.
  \end{itemize}
In the sequel we will use either random classification, $k$-nearest neighbors (with different distances), or maximum Mahalanobis depth (with moment and robust estimates).

\section{The $\alpha$-procedure}\label{alpha}

To separate the $q$ classes in the multi-depth space we use the \textit{$\alpha$-procedure}, which has been developed by Vasil'ev \cite{Vasilev91,Vasilev03} and Lange \cite{VasilevL98}, see also \cite{LangeMB11}.
Among others the regression depth method (see \cite{RousseeuwH99,ChristmannR01} or \cite{ChristmannFJ02}) or the support vector machine (see \cite{Vapnik98} and \cite{ChristmannFJ02}) seem to be good alternatives. In contrast with those the $\alpha$-procedure, in application to the current task, is substantially faster and produces a  unique decision rule. Besides that it focuses on features of the extended $[0,1]^q$, i.e. depths and their products, which, by their nature, are rather relevant. Moreover, by selecting a few important features only, the $\alpha$-procedure yields a rather stable solution.


Let us first present the procedure in the case of $q=2$ classes. As above consider two clouds of training data in $\mathbb{R}^d$, $X=\{{\bf x}_1,\dots,{\bf x}_{n_1}\}$ and $Y=\{{\bf y}_1,\dots,{\bf y}_{n_2}\}$ and notate ${\bf x}_{n_1+m}={\bf y}_m$, $m=1,...,n_2$. By calculating the depth of all ${\bf x}_i$ with respect to each of the two clouds, their depth representation, $(D_X({\bf x}_i),D_Y({\bf x}_i))$, is obtained, $i=1,2,\dots, n_1+n_2$. The set
   \[{\cal D}=\{ {\bf d}_i\in [0,1]^2|{\bf d}_i=(D_X({\bf x}_i),D_Y({\bf x}_i)),i=1,\dots ,n_1+n_2\}\]
is the DD-plot of the data (\cite{LiuPS99}).

We use a modified version of the $\alpha$-procedure to construct a nonlinear separator in $[0,1]^2$ that classifies the D-represented data points.
The construction is based on depth values and the products of depth values up to some degree $p$ that can be either chosen \textit{a priori} or determined by cross-validation.
For this, a linearized representation of the two classes in a depth feature space is 
\begin{eqnarray*}
  {\bf Z}= \{{\bf  z}_i &|& {\bf  z}_i=\left(D_X({\bf x}_i),D_Y({\bf x}_i),D_X({\bf x}_i)\cdot D_Y({\bf x}_i),D^2_X({\bf x}_i),D^2_Y({\bf x}_i)\right),\\
  && i=1,...,n_1+n_2\}\,.
\end{eqnarray*}

Each element of the extended D-representation is mentioned as a \textit{basic D-feature} and the space $[0,1]^r$ as the \textbf{feature space}.
When the maximum exponent is $p\ge 1$, ${\bf z}_i$ is a vector in $\mathbb{R}^r$ having components
\begin{equation}\label{D-feature}
    D_X({\bf x}_i)^{k_\nu} \cdot D_Y({\bf x}_i)^{\ell_\nu}, \quad \text{where}\;\; 1\le k_\nu+\ell_\nu \le p\,, \quad \nu=1,\dots,r\,.
\end{equation}
The number of basic D-features, that is the dimension of the feature space, equals $r = {p+2 \choose 2}-1$, which is easily seen by induction.
We index the basic D-features by $\nu$ and notate ${\bf z}_i = (z_{i\nu})_{\nu=1,\dots,r}.$

The $\alpha$-procedure now, in a stepwise way, performs linear discrimination in subspaces of the feature space. It is a bottom-up approach that successively builds new features from the basic D-features. In each step certain two-dimensional subspaces of $\bf Z$ are considered, and the projection of ${\bf Z}$ to each of these subspaces is separated by a straight discrimination line.
Out of these subspaces the $\alpha$-procedure selects a subspace whose discrimination line provides the least classification error.
Clearly any discrimination line that separates the DD-plot must pass through the origin since $D_X({\bf x}_i)=D_Y({\bf x}_i)=0$ implies that the point ${\bf x}_i$ cannot be classified to either of the two classes. The same must hold for all discrimination lines  in subspaces of the extended depth space.

\begin{center}
\begin{figure}[h!]
  \includegraphics[keepaspectratio=true,scale=0.90]{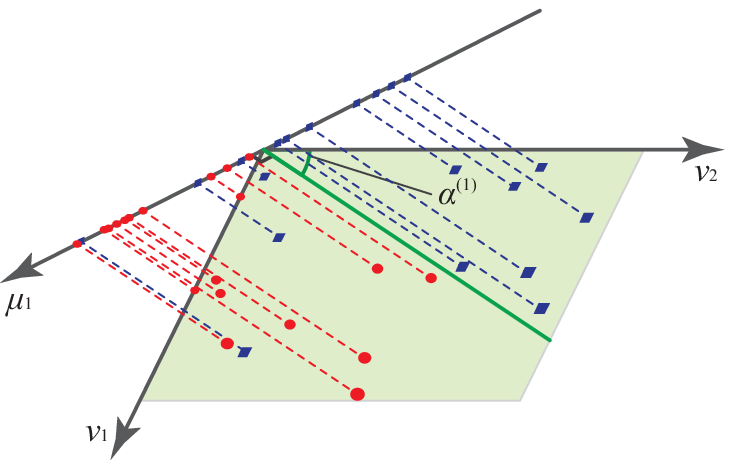}
  \caption{$\alpha$-procedure; step 1.}
  \label{alphaFigure1}
\end{figure}
\end{center}

In a \textit{first step} a pair $(\nu_1,\nu_2)$ of D-features (\ref{D-feature}) is chosen with $(k_1+k_2)(\ell_1+\ell_2)>0$. The latter restriction implies that the two D-features do not solely relate to one of the classes. A straight discrimination line is calculated in the two-dimensional coordinate subspace defined by the pair $(\nu_1,\nu_2)$. As the line passes through the origin it is characterized by an angle $\alpha\in [0,2 \pi[$. The best discriminating angle $\alpha_{\nu_1,\nu_2}$ is determined by minimizing the \textit{average misclassification rate (AMR)},
\begin{eqnarray}\label{ACE}
  \Delta(\alpha; \nu_1,\nu_2) &=& \frac{1}{n_1+n_2} \Big[\sum_{i=1}^{n_1} I({z}_{i,\nu_1}\cos \alpha + {z}_{i,\nu_2}\sin \alpha < 0)    \\
 \nonumber   &&  \qquad  \quad \;\; + \sum_{i=n_1+1}^{n_1+n_2} I({z}_{i,\nu_1}\cos \alpha + {z}_{i,\nu_2}\sin\alpha > 0)\Big]\,.
\end{eqnarray}
Here $I(A)$ denotes the indicator function of $A$. If the minimum is attained in an interval, its middle value is selected for $\alpha_{\nu_1,\nu_2}$; see Figure \ref{alphaFigure1}.
The same is done for all pairs of D-features satisfying the above restriction, and the pair $(\nu^*_1,\nu^*_2)$ is selected that minimizes (\ref{ACE}). If the minimum is not unique the pair with the smallest $k$ and $\ell$ is chosen.
Let $\alpha^{(1)}= \alpha_{\nu^*_1,\nu^*_2}$ and denote the respective AMR by $\Delta^{(1)}$. Next the D-features $\nu^*_1$ and $\nu^*_2$ are replaced by a new D-feature which is indexed by $\mu_1$ and gives value
\begin{equation}\label{D-feature(1)}
   z_{i,\mu_1}= z_{i,\nu_1}\cos \alpha^{(1)} + z_{i,\nu_2}\sin \alpha^{(1)}\,, \quad i=1,\dots n_1+n_2\,,
\end{equation}
to each ${\bf x}_i$. Geometrically the values are obtained by projecting $(z_{i,\nu_1},z_{i,\nu_2})$ to a straight line in the $({\nu_1},{\nu_2})$-plane that is perpendicular to the discrimination line; see Figure \ref{alphaFigure1}.
The first step results in the new D-feature $\mu_1$ and the AMR $\Delta^{(1)}$ produced by classifying according to this feature.

\begin{center}
\begin{figure}
  \includegraphics[keepaspectratio=true,scale=0.90]{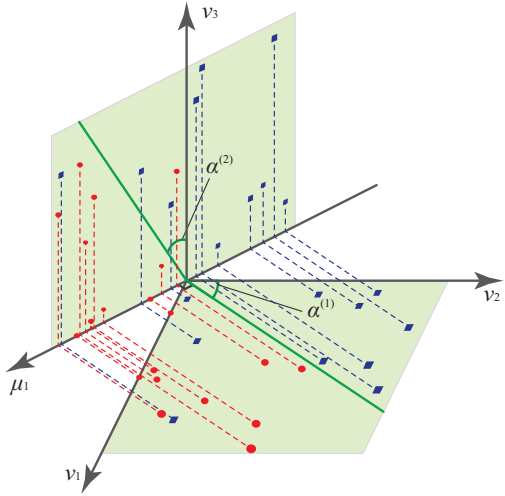}
  \caption{$\alpha$-procedure; step 2.}
  \label{alphaFigure2}
\end{figure}
\end{center}

The \textit{second step} couples the new D-feature $\mu_1$ with each of the basic D-features $\nu$ that have not been replaced so far. For each of these pairs of D-features a best discriminating angle $\alpha_{\mu_1,\nu}$ is determined, and among these the pair of D-features is selected that provides the minimum AMR. The minimum error is denoted by $\Delta^{(2)}$ and the angle at which it is attained by $\alpha^{(2)}$.
This is visualized in Figure~\ref{alphaFigure2}.
The best pair of D-features is replaced by a new D-feature $\mu_2$, where the values $z_{i,\mu_2}$ are calculated as in (\ref{D-feature(1)}).

The last step is repeated with $\mu_2$ in place of $\mu_1$, etc. The procedure stops after step $t$ if either the additional {discriminating power} $\Delta^{(t)}-\Delta^{(t+1)}=0$
or $t=r$, that is, all basic D-features have been replaced. Then the angle $\alpha^{(t)}$ defines a linear rule for discriminating between two (up to) $p$-th order polynomials in $D_X(\bf z)$ and $D_Y(\bf z)$, which correspond to the two finally constructed D-features, according to their sign. This yields a polynomial separation of the classes in the depth space.

For example, let in step 1 the basic features $D_X$ and $D_Y^2$ be selected and, consequently, $D_X\cdot D_Y$ and $D_X^2$ be included in steps 2 and 3. If the procedure terminates after step 3, the result is a polynomial in the two depths $D_X({\bf x})$ and $D_Y({\bf x})$ that has form
\[ a D_X({\bf x}) + b D_X^2({\bf x}) + c D_Y^2({\bf x}) + d D_X({\bf x}) D_Y({\bf x})\]
A given point ${\bf x}$ of the object space then is classified according to the sign of the polynomial.

If there are more than two classes, say $X_1, \dots, X_q$, each data point ${\bf x}_i$ is represented by the vector of depth values ${\bf d} = (D_{X_1}({\bf x}_i),\dots, D_{X_q}({\bf x}_i))$ in $[0,1]^q$. Again a depth feature space is considered of some order $p$; it has dimension $r={p+q \choose q}-1$.
With $q>2$ classes every two training classes $X_j, X_k, j\not= k,$ are separated by the $\alpha$-procedure in the same way as above: In each step a pair of D-features is replaced by a new D-feature as long as the AMR decreases and basic D-features are left to be replaced.
For each pair of classes the procedure results in a hypersurface that separates the $q$-dimensional depth space into two sets of attraction. A given point ${\bf x}$ is finally assigned to that class to which it has been most often attracted.

\section{Some theoretical aspects}\label{props}

In order to investigate some properties of the DD$\alpha$-approach we transfer it to a more general probabilistic setting and define a depth function as the population version of a data depth. Let ${\cal P}$ be  a properly chosen set of probability distributions on $\IR^d$ that includes the empirical distributions. A \textit{depth function} $D$ is a function that assigns a value $D_P(\bf x)\in [0,1]$ to every ${\bf x}\in \IR^d$ and $P\in{\cal P}$ in an affine invariant way (i.e. $D_{AP+b}(A{\bf x}+b)=D_P({\bf x})$ for any nonsingular matrix $A\in \IR^{d\times d}$ and any $b\in \IR^d$, $AP$ denoting the push-forward measure), and has convex compact upper level sets.
Obviously, the restriction of a depth function $D$ to the class of empirical distributions is an affine invariant quasiconvex data depth.
For details on general depth functions, see e.g.\ the above cited surveys \cite{Cascos09,Mosler02,Serfling06,ZuoS00}.

While data depth is an intrinsically nonparametric notion, the behavior of depth functions and depth based procedures on parametric classes is of special interest as it indicates how the nonparametric approach relates to the more classical parametric one. As a generalization of multivariate Gaussian distributions, spherical and elliptical distributions play an important role in parametric multivariate analysis.
A random vector $\bf X$ in $\IR^d$ has a \textit{spherical distribution} if ${\bf X}= R\cdot \bf U$, where $\bf U$ is a random vector uniformly distributed  on the sphere $S^{d-1}$ and $R$ is a random variable having support $[0,\infty[$ and being independent of $\bf U$. A random vector $\bf Y$ has an \textit{elliptical distribution} if it is an affine transform of a spherically distributed $\bf X$, ${\bf Y} = \mu + B {\bf X}$. If $R$ has a density $r$ we notate
${\bf Y}\sim {\rm Ell}(\mu, BB', r)$. As, by definition, a depth function is affine invariant, it operates on elliptical distributions in a rather simple way.
The following propositions give some insight into the the behavior of depth functions and the DD$\alpha$-procedure if the data generating processes are elliptical.

\begin{proposition}
    If $D$ is an affine invariant depth function and $P$ an elliptical distribution, then for every $\alpha\in ]0,1]$ the upper level set
    \[D_\alpha(P)=\{{\bf x}\in \IR^d | D_P({\bf x})\ge \alpha\}\]
    is an ellipsoid.
\end{proposition}

\textbf{Proof.} Let $P={\rm Ell}(\mu, BB', r)$ and $\alpha\in ]0,1]$.
Consider $P_0={\rm Ell}(\vec 0, {I}_d, r)$. Then, for all $\beta\ge \alpha$,  $\{{\bf x}\in \IR^d | D_{P_0}({\bf x})=\beta\}$
is a sphere since $D$ is, in particular, orthogonal invariant. Hence, $D_\alpha(P_0)=\{{\bf x}\in \IR^d | D_{P_0}({\bf x})\ge \alpha\}$ is a ball and, by affine transformation with $\mu$ and $B$, $D_\alpha(P)$ is an ellipsoid. \hfill $\Box$

\begin{proposition}\label{prop2}
 \begin{description}
  \item[(i)]   Let $D$ be the zonoid depth and $P$ a unimodal elliptical distribution, that is $P={\rm Ell}(\mu, BB', r)$. Then, for every non-empty density level set
   $\{{\bf x}\in \IR^d | f({\bf x})\ge \beta\}$, some $\alpha=\phi(\beta)$ exists such that
   \[\{{\bf x}\in \IR^d | f({\bf x})\ge \beta\}=D_\alpha(P)\,.\]
  \item[(ii)] If, in addition, $r$ has an interval support
  then $\phi$ is a continuous, strictly increasing function.
  It holds $D_P({\bf x})=\phi(f({\bf x}))$ and therefore
  \begin{equation}\label{equivalentrule}
f({\bf x})\ge f({\bf y}) \quad \Longleftrightarrow \quad D_P({\bf x}) \ge D_P({\bf y})\,.
  \end{equation}

 \end{description}
\end{proposition}

\textbf{Proof.} (i): Note that $D_0=\mathbb{R}^d$. Thus, if $\beta \le 0$, the claim holds with $\alpha=0$. Now let $\beta>0$ and assume w.l.o.g. that $P$ is spherical. Then $\{{\bf x}\in \IR^d | f({\bf x})\ge \beta\}$ is a ball with center at the origin. Let ${\bf x}^*$ be a point on its surface.
Also the central regions $D_\alpha$ are balls around the origin. By Theorems 3.9 and 3.14 in \cite{Mosler02}, the $D_\alpha$ are continuous and strictly decreasing on the convex hull of the support of $P$ and it holds $\alpha^*:=D_P({\bf x}^*)>0$.  We conclude $D_{\alpha^*}=\{{\bf x}\in \IR^d | f({\bf x})\ge \beta\}$.\\
(ii): Under the additional premise, the density level sets are continuously and strictly decreasing in $\beta>0$, which yields the result.
\hfill $\Box$

\begin{corollary}
Consider a mixture of unimodal elliptical distributions $P_j={\rm Ell}(\mu_j, B_jB_j^\prime, r_j)$, $j=1,\dots, q$, with
mixing probabilities $\pi_j$
and assume that all $r_j$ have an interval support.
Let $D$ be the zonoid depth.

Then, for each $j$ and $k$ exists a strictly increasing function $\psi_{jk}$ so that
\[ \pi_j \cdot f_j({\bf x}) < \pi_k \cdot f_k({\bf x})    \quad \Longleftrightarrow \quad D_{P_j}({\bf x})< \psi_{jk} (D_{P_k}({\bf x}))\,.
\]
\end{corollary}

\textbf{Proof.} From Proposition \ref{prop2} continuous and strictly increasing functions $\phi_j$ and $\phi_k$ are obtained with
$D_{P_j}({\bf x})= \phi_j(f_j({\bf x}))$ and $D_{P_k}({\bf x})= \phi_k(f_k({\bf x}))$. Consequently,
\[ \pi_j \cdot f_j({\bf x}) < \pi_k \cdot f_k({\bf x}) \quad \Leftrightarrow \quad D_{X_j}({\bf x}) < \phi_j \left(\frac{\pi_k}{\pi_j} \, \phi_k^{-1}(D_{X_k}({\bf x}))\right)\,,
\]
which proves the claim by use of the function $\psi_{jk}(\cdot) = \phi_j \left(\frac{\pi_k}{\pi_j} \,\phi_k^{-1}(\cdot)\right)$\,.
\hfill $\Box$

A similar result holds for other data depths including the halfspace, simplicial, projection and Mahalanobis depths; see Prop. 1 in \cite{LiCAL11}. In the rest of section we consider the limit behavior of the DD$\alpha$-procedure under independent sampling. For this, we assume that the empirical depth is a consistent estimator of its population version. This is particularly true for the zonoid, halfspace, simplicial, projection and Mahalanobis depths.

\begin{theorem}[Bayes rule]
Let $F$ and $G$  probability distributions in $\IR^d$ having densities $f$ and $g$, and let $H$ be a hyperplane such that $G$ is the mirror image of $F$ with respect to $H$ and $f\ge g$ in one of the half-spaces generated by $H$. Then based on a 50:50 independent sample from $F$ and $G$ the DD$\alpha$-procedure will asymptotically yield the linear separator that corresponds to the bisecting line of the DD-plot.
\end{theorem}
Note that the rule given in the theorem corresponds the Bayes rule, see \cite{HastieTF09}. Especially the requirements of the theorem are satisfied if $F$ and $G$ are mirror symmetric and unimodal.

\textbf{Proof.} Due to the mirror symmetry of the distributions in $\IR^d$ the DD-plot is symmetric as well. Symmetry axis is the bisector, which is obviously  the result of the $\alpha$-procedure when the sample is large enough. \hfill $\Box$

\begin{theorem}
    Let $F, G$ be unimodal elliptical, $F={\rm Ell}(\mu_F, BB^\prime, r)$, $G=$ ${\rm Ell}(\mu_G, BB^\prime, r)$. Then based on a 50:50 independent sample from $F$ and $G$ the DD$\alpha$-procedure will asymptotically yield the linear separator that corresponds to the bisecting line of the DD-plot.
\end{theorem}

\textbf{Proof.} If $F$ and $G$ are spherically symmetric, they satisfy the premise of the previous theorem. A common affine transformation of $F$ and $G$ does not change the DD-plot. \hfill $\Box$

\section{Simulation study}\label{comps}

The DD$\alpha$-procedure has been implemented on a standard PC in an $R$-environ\-ment.
To explore its specific potencies we apply it to simulated as well as to real data. The same data have been analyzed with several classifiers in the literature. In this section results on simulated data are presented regarding the average misclassification rate of nine procedures besides the DD$\alpha$-classifier (Section \ref {compsPerfm}). Then the speed of the DD$\alpha$-procedure is quantified (Section~\ref{compsSpeed}). The following Section \ref{secbenchmark}
covers the relative performance of the the DD$\alpha$- and other classifiers on several benchmark data sets.

\subsection{Comparison of performance}\label{compsPerfm}

To simplify the comparison with known classifiers, we use the same simulation settings as in \cite{LiCAL11}.
These are supervised classification tasks with two equally sized training classes. Data are generated by ten pairs of distributions according to Table~\ref{distributionsTable}. Here N and Exp denote the Gaussian and exponentional distributions, respectively, and \\
 MixN$(\mu, \sigma_1, \sigma_2)=\begin{cases}
    -\sigma_1*\left|\text{N}(0,1)\right|+\mu & \text{with probability }1/2,\\
    \sigma_2*\left|\text{N}(0,1)\right|+\mu & \text{with probability }1/2.
 \end{cases}$\\

\begin{table}
\caption{Distributional settings used in the simulation study.}
\label{distributionsTable}
\begin{tabular}{|c|c|l|l|}
\hline No. & Alternative & 1st class & 2nd class \\
\hline \hline 1 & Normal &
\multirow{2}{*}{N$(\bigl[\begin{smallmatrix} 0\\0 \end{smallmatrix}\bigr],\bigl[\begin{smallmatrix} 1 & 1\\ 1 & 4 \end{smallmatrix}\bigr])$} &
\multirow{2}{*}{N$(\bigl[\begin{smallmatrix} 1\\1 \end{smallmatrix}\bigr],\bigl[\begin{smallmatrix} 1 & 1\\ 1 & 4 \end{smallmatrix}\bigr])$} \\
 & location & & \\

\hline 2 & Normal &
\multirow{2}{*}{N$(\bigl[\begin{smallmatrix} 0\\0 \end{smallmatrix}\bigr],\bigl[\begin{smallmatrix} 1 & 1\\ 1 & 4 \end{smallmatrix}\bigr])$} &
\multirow{2}{*}{N$(\bigl[\begin{smallmatrix} 1\\1 \end{smallmatrix}\bigr],\bigl[\begin{smallmatrix} 4 & 4\\ 4 & 16 \end{smallmatrix}\bigr])$} \\
 & location-scale & & \\

\hline 3 & Cauchy &
\multirow{2}{*}{Cauchy$(\bigl[\begin{smallmatrix} 0\\0 \end{smallmatrix}\bigr],\bigl[\begin{smallmatrix} 1 & 1\\ 1 & 4 \end{smallmatrix}\bigr])$} &
\multirow{2}{*}{Cauchy$(\bigl[\begin{smallmatrix} 1\\1 \end{smallmatrix}\bigr],\bigl[\begin{smallmatrix} 1 & 1\\ 1 & 4 \end{smallmatrix}\bigr])$} \\
 & location & & \\

\hline 4 & Cauchy &
\multirow{2}{*}{Cauchy$(\bigl[\begin{smallmatrix} 0\\0 \end{smallmatrix}\bigr],\bigl[\begin{smallmatrix} 1 & 1\\ 1 & 4 \end{smallmatrix}\bigr])$} &
\multirow{2}{*}{Cauchy$(\bigl[\begin{smallmatrix} 1\\1 \end{smallmatrix}\bigr],\bigl[\begin{smallmatrix} 4 & 4\\ 4 & 16 \end{smallmatrix}\bigr])$} \\
 & location-scale & & \\

\hline 5 & Normal & Learning sample: 90\% as No. 1, & as No. 1 \\
 & contaminated & 10\% from N$(\bigl[\begin{smallmatrix} 10\\10 \end{smallmatrix}\bigr],\bigl[\begin{smallmatrix} 1 & 1\\ 1 & 4 \end{smallmatrix}\bigr])$. & \\
 & location & Testing sample: as No. 1 & \\

\hline 6 & Normal & Learning sample: 90\% as No. 2, & as No. 2 \\
 & contaminated & 10\% from N$(\bigl[\begin{smallmatrix} 10\\10 \end{smallmatrix}\bigr],\bigl[\begin{smallmatrix} 1 & 1\\ 1 & 4 \end{smallmatrix}\bigr])$. & \\
 & location-scale & Testing sample: as No. 2 & \\

\hline 7 & Exponential &
\multirow{2}{*}{$(\text{Exp}(1),\text{Exp}(1))$} &
\multirow{2}{*}{$(\text{Exp}(1) + 1,\text{Exp}(1) + 1)$} \\
 & location & & \\

\hline 8 & Exponential &
\multirow{2}{*}{$(\text{Exp}(1),\text{Exp}(1/2))$} &
\multirow{2}{*}{$(\text{Exp}(1/2) + 1,\text{Exp}(1) + 1)$} \\
 & location-scale & & \\

\hline 9 & Asymmetric &
\multirow{2}{*}{$(\text{MixN}(0; 1, 2),\text{MixN}(0; 1, 4))$} &
\multirow{2}{*}{$(\text{MixN}(1; 1, 2), \text{MixN}(1; 1, 4))$} \\
 & location & & \\

\hline 10 & Normal- &
\multirow{2}{*}{N$(\bigl[\begin{smallmatrix} 0\\0 \end{smallmatrix}\bigr],\bigl[\begin{smallmatrix} 1 & 0\\ 0 & 1 \end{smallmatrix}\bigr])$} &
\multirow{2}{*}{$(\text{Exp}(1),\text{Exp}(1))$} \\
 & exponential & & \\

\hline
\end{tabular}
\end{table}

The DD$\alpha$-classifier is contrasted with the following nine classifiers: linear discriminant analysis (LDA), quadratic discriminant analysis (QDA), $k$-nearest neighbors classification ($k$-NN), maximum depth classification based on Mahalanobis (MM), simplicial (MS), and halfspace (MH) depth, and DD-classification with the same depths (DM, DS and DH, correspondingly).
For more details about the data and the procedures as well as for some motivation the reader is referred to \cite{LiCAL11}.

All simulations of \cite{LiCAL11} are recalculated following their paper as close as possible. The LDA, QDA and $k$-NN classifiers are computed with the R-packages ``MASS'' and ``class'', where the parameter $k$ of the $k$-NN-classifier is selected by leave-one-out cross-validation over a relatively wide range.
 The simplicial, and halfspace depths
have been determined by exact calculations with the R-package ``depth''.
The zonoid depth has been exactly computed by the algorithm in \cite{DyckerhoffKM96}.
Recall that, in dimension two, calculations of all these depths can be efficiently done by a circular sequence
and note that the problem of prior probabilities
is avoided by choosing test samples of equal size from both classes.

\begin{figure*}
  \includegraphics[keepaspectratio=true,scale=0.40]{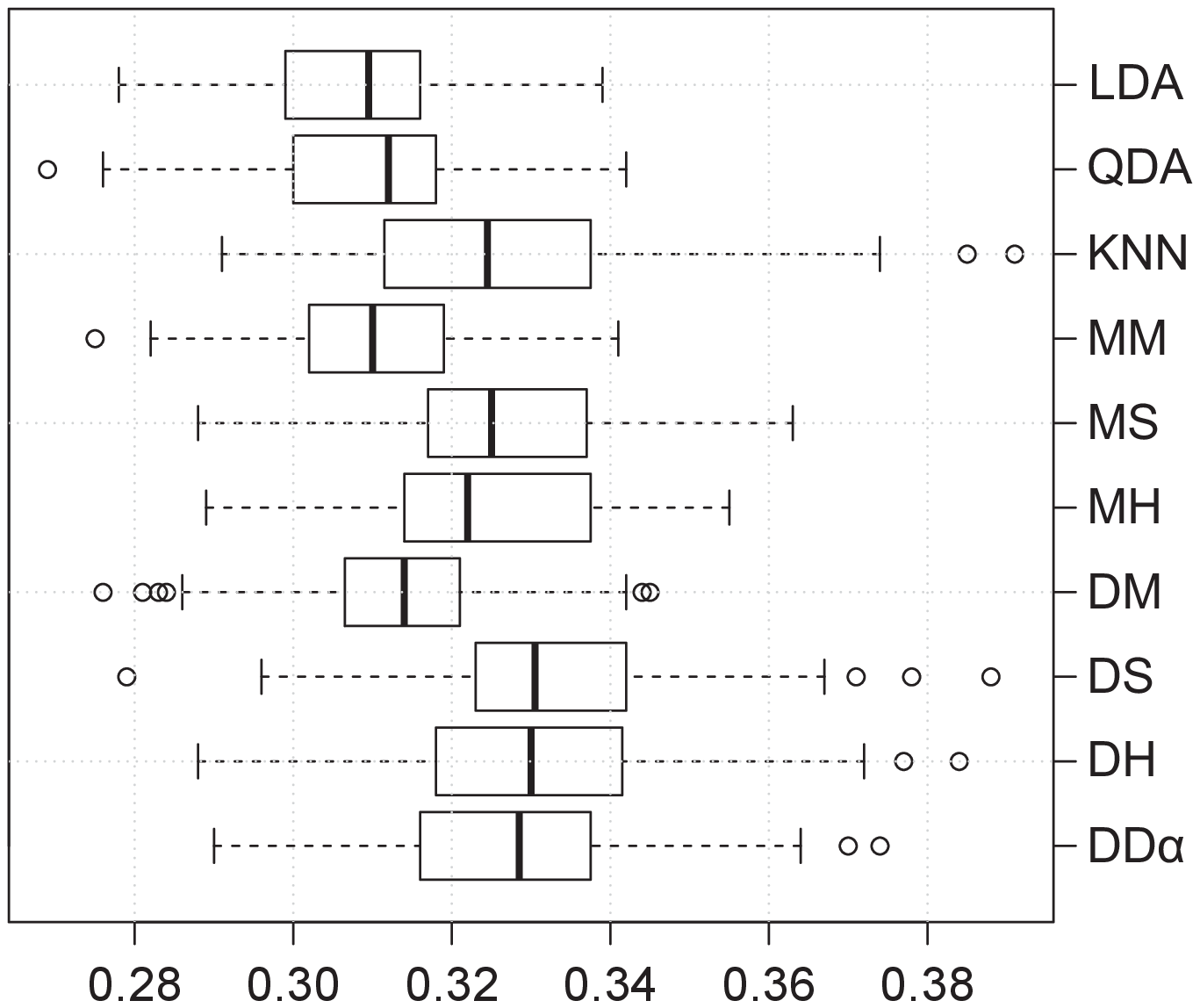}
  \includegraphics[keepaspectratio=true,scale=0.40]{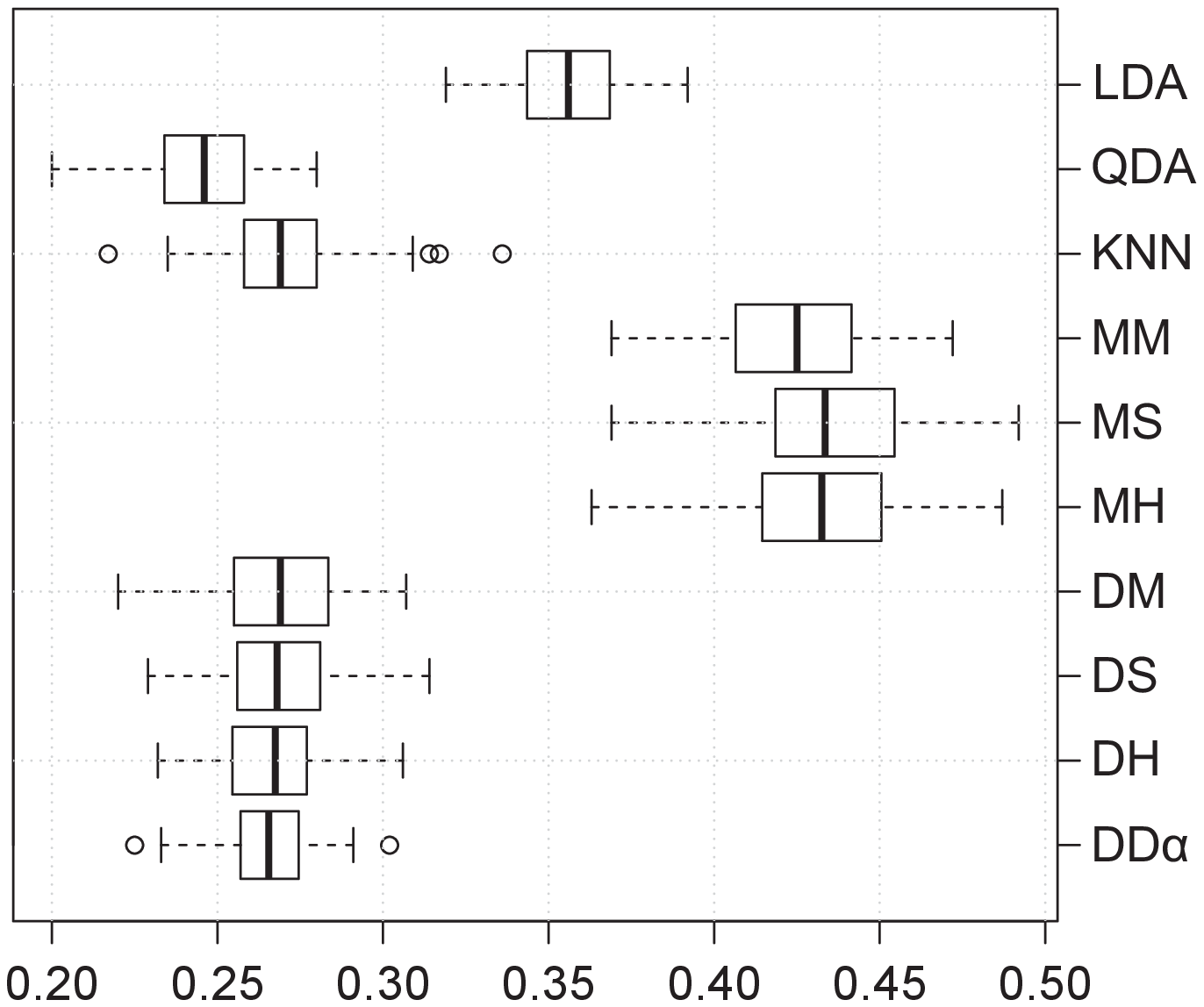}
  \caption{Normal location (left) and location-scale (right) alternatives.}
  \label{simulationsFigure12}
\end{figure*}

\begin{figure*}
  \includegraphics[keepaspectratio=true,scale=0.40]{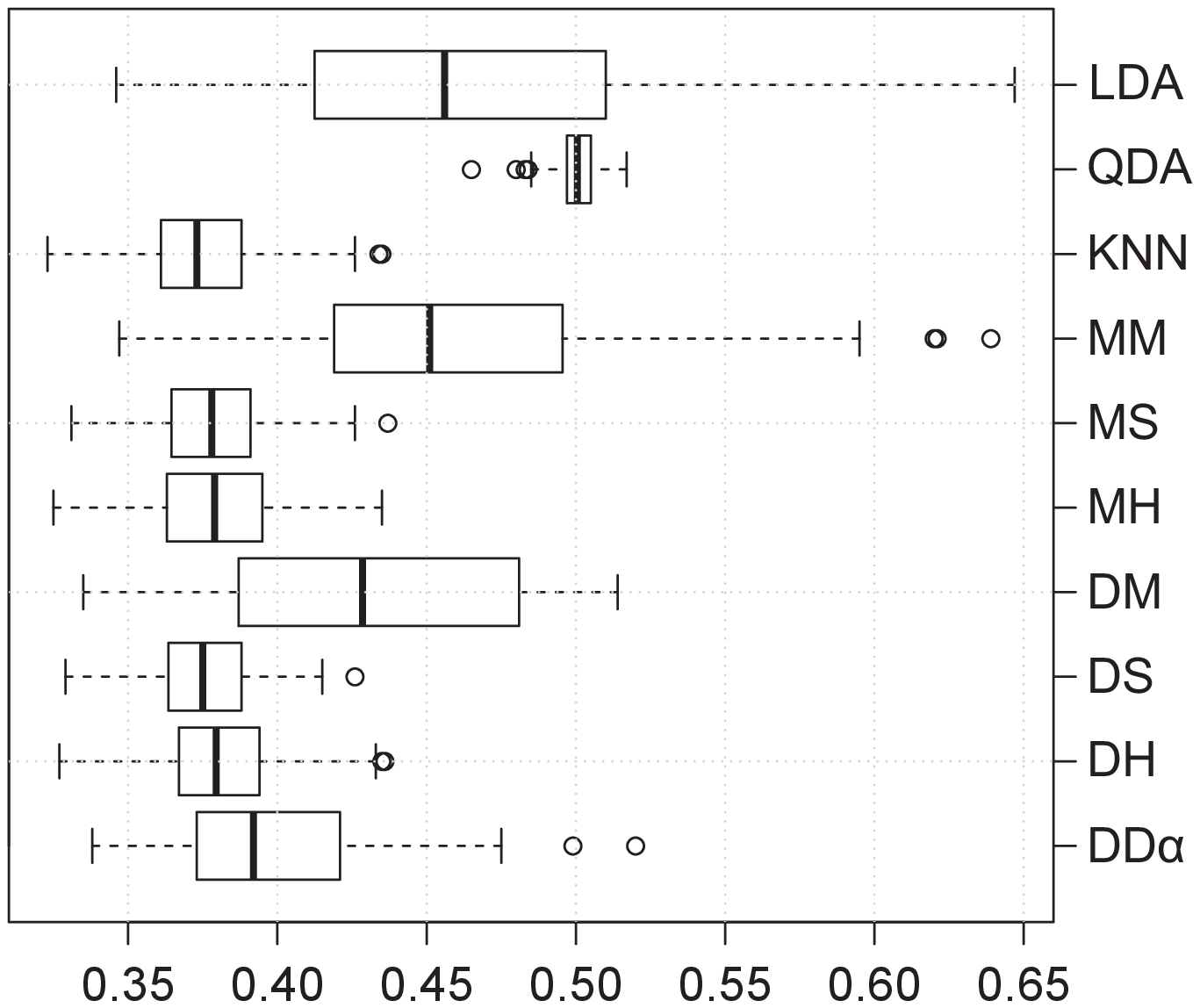}
  \includegraphics[keepaspectratio=true,scale=0.40]{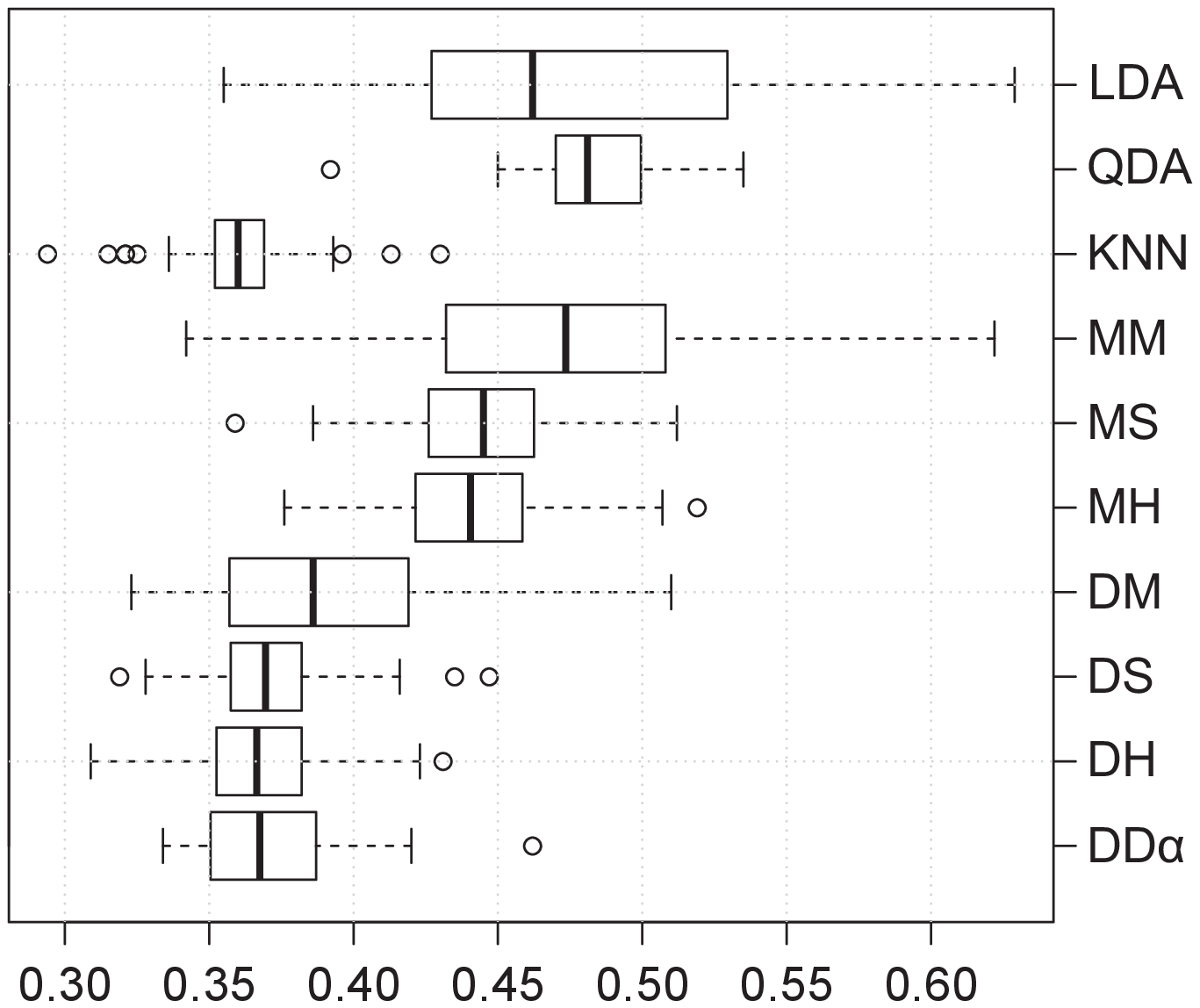}
  \caption{Cauchy location (left) and location-scale (right) alternatives.}
  \label{simulationsFigure34}
\end{figure*}

\begin{figure*}
  \includegraphics[keepaspectratio=true,scale=0.40]{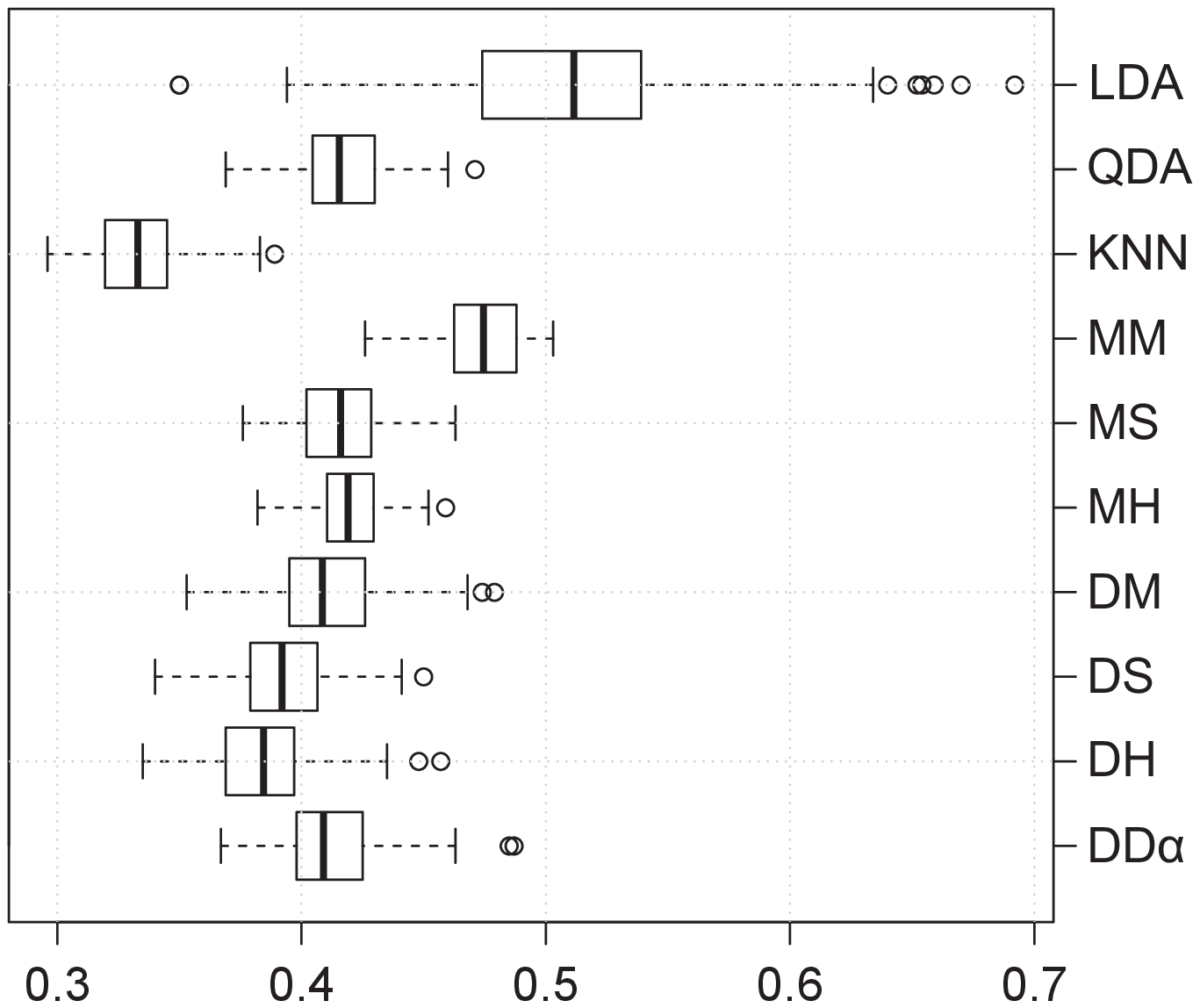}
  \includegraphics[keepaspectratio=true,scale=0.40]{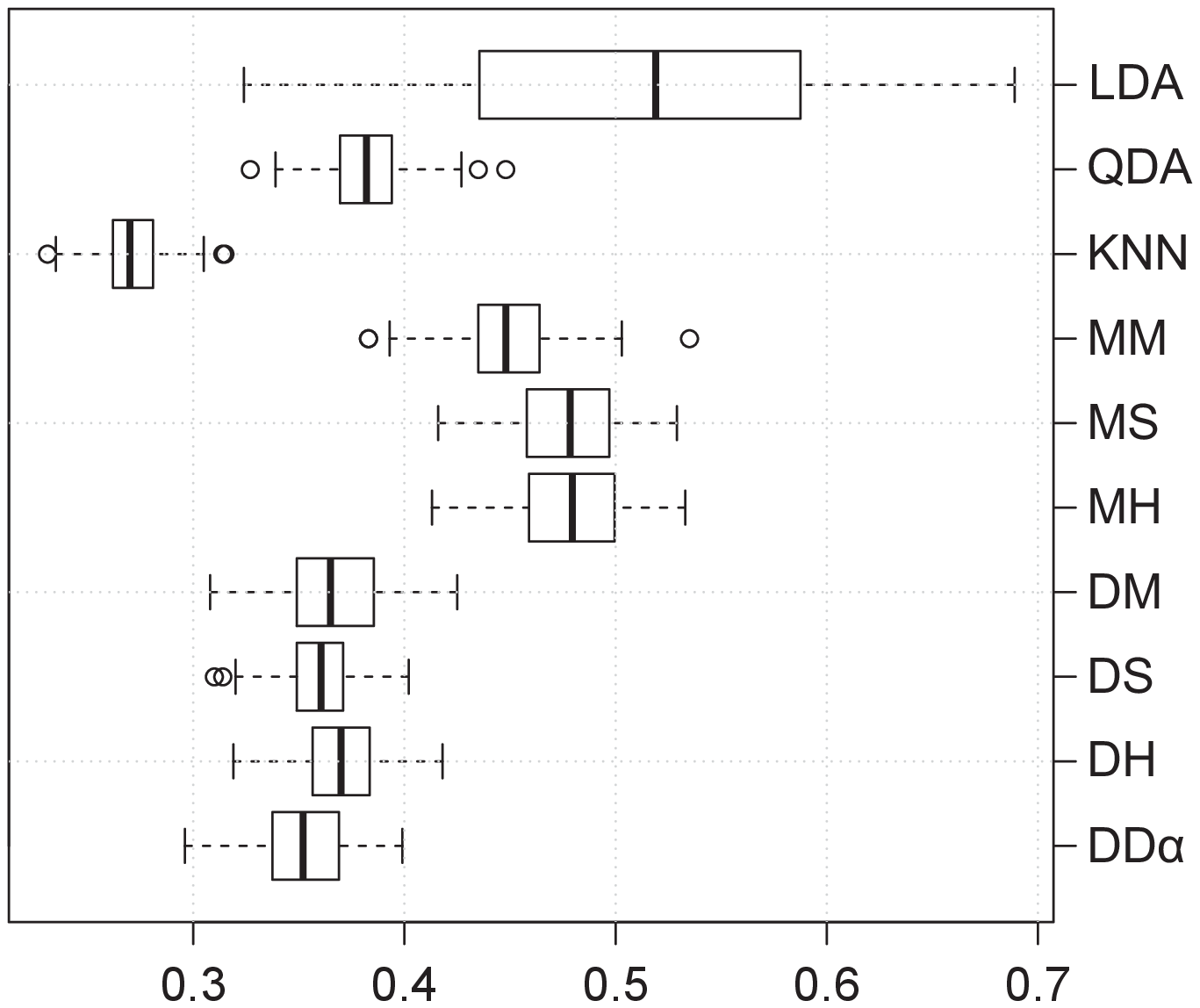}
  \caption{Normal contaminated location (left) and location-scale (right) alternatives.}
  \label{simulationsFigure56}
\end{figure*}

\begin{figure*}
  \includegraphics[keepaspectratio=true,scale=0.40]{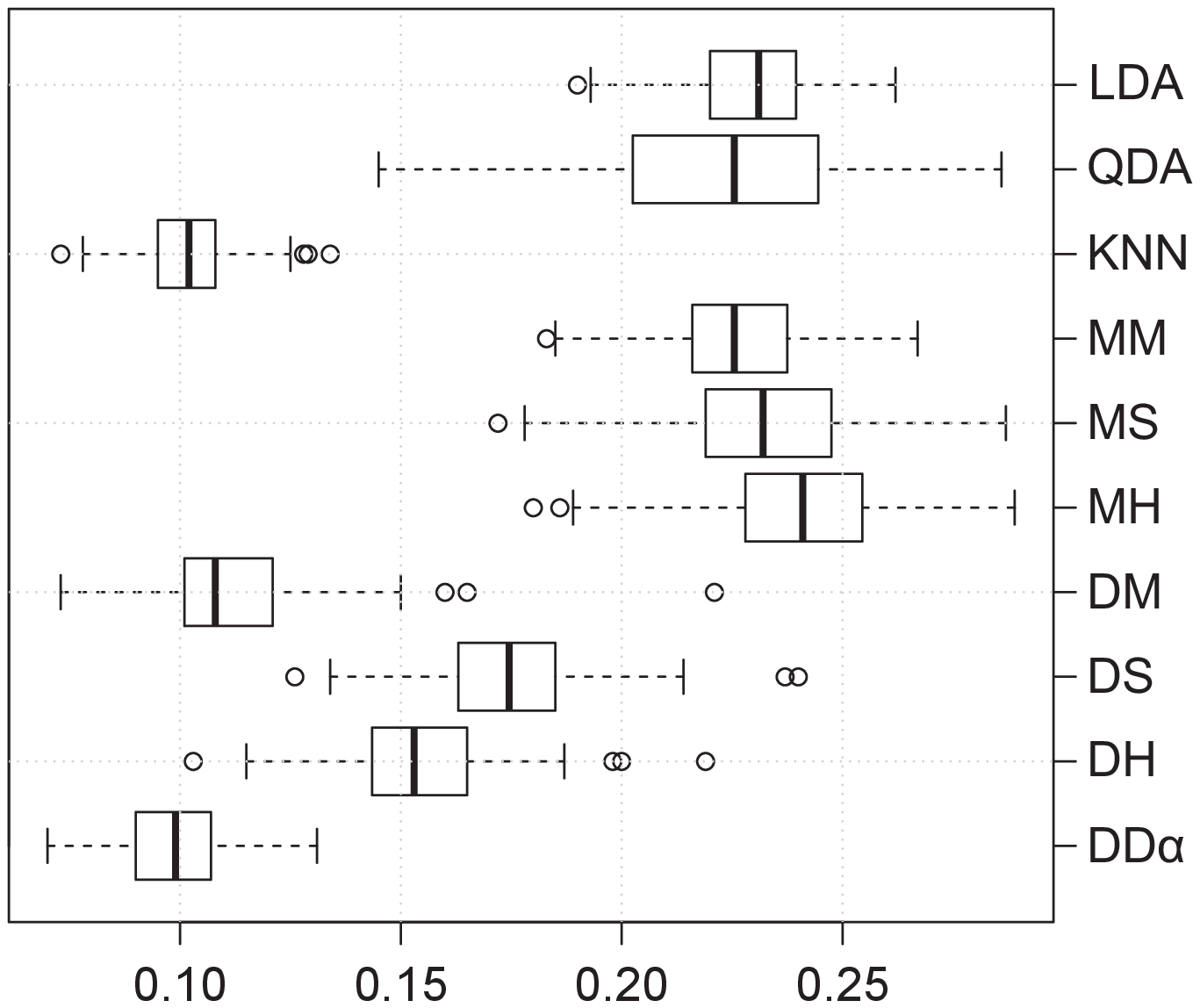}
  \includegraphics[keepaspectratio=true,scale=0.40]{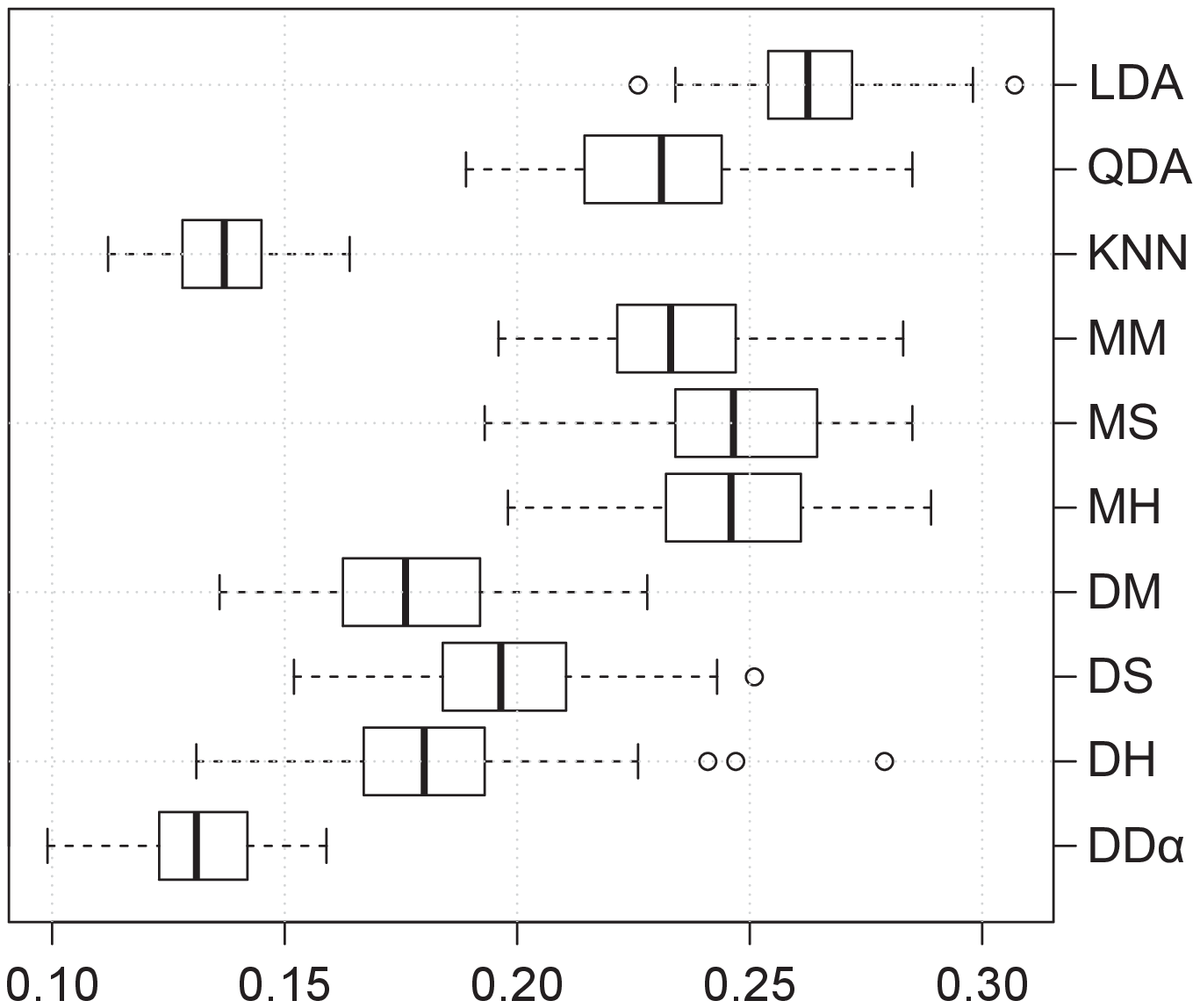}
  \caption{Exponential location (left) and location-scale (right) alternatives.}
  \label{simulationsFigure78}
\end{figure*}

\begin{figure*}
  \includegraphics[keepaspectratio=true,scale=0.40]{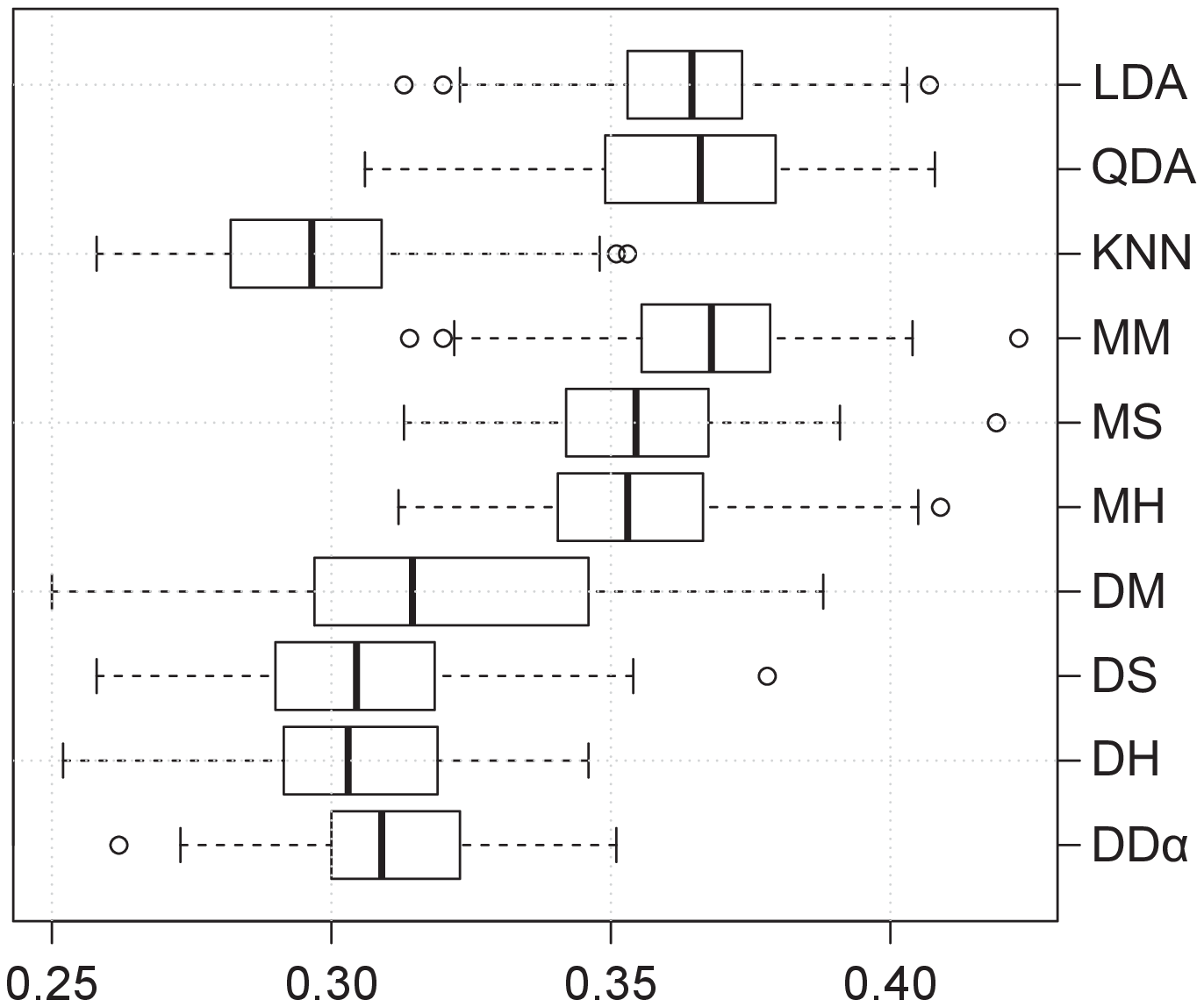}
  \includegraphics[keepaspectratio=true,scale=0.40]{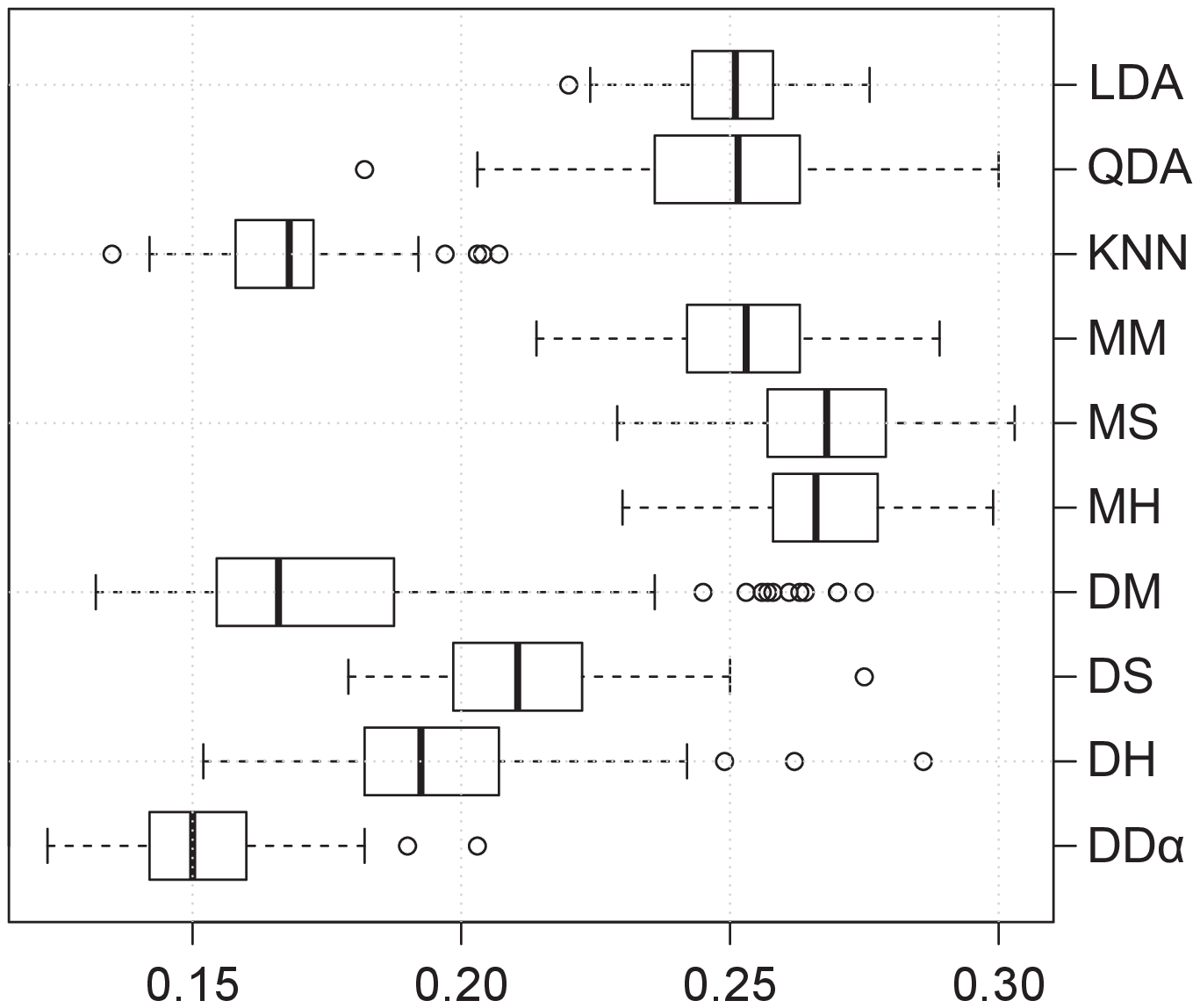}
  \caption{Asymmetric location (left) and normal-exponential (right) alternatives.}
  \label{simulationsFigure910}
\end{figure*}

For the DD-classifiers a polynomial line (up to degree three) is determined to discriminate in the two-dimensional DD-Plot, a tenfold cross-validation is employed to choose the optimal degree of the polynomial, a smoothing constant $t=100$ is selected in the logistic function, and the DD-Plot is never rotated.
Each experiment includes a training phase and an evaluation phase:
From the given pair of distributions 400 observations (200 of each class) are generated to train the classifier, and 1000 (500 of each) observations to evaluate its AMR.
For each distribution pair and each classifier 100 experiments  are performed, and the resulting sample of AMRs is visualized as a box-plot; see Figures \ref{simulationsFigure12} to \ref{simulationsFigure910}.

As we have discussed at the end of Section \ref{ddplot},
with depths like the simplicial, halfspace and zonoid depth the problem of outsiders arises. An outsider is, in the DD-plot, represented by the origin.
A simple approach is to assign the outsiders randomly to the two classes.
Throughout our simulation study we have chosen the random assignment rule, which results in kind of worst case AMR.
Observe that this choice of assignment rule discriminates against the procedures that yield outsiders and advantages those that do not, in particular
LDA, QDA, MM, DM and $k$-NN for all distribution settings.

The principal results of the simulation study are collected in Figures \ref{simulationsFigure12} to \ref{simulationsFigure910}.
Under the normal location-shift model (Figure~\ref{simulationsFigure12}, left) all classifiers behave satisfactorily, and the DD$\alpha$-classifier performs well among them.
However LDA, QDA, MM and DM show slightly better results since they do not have to cope with outsiders like the other depth-based procedures.

Also under the normal location-scale alternative (Figure~\ref{simulationsFigure12}, right) the DD$\alpha$-classifier performs rather well, like all DD-classifiers.
A slightly worse performance of the DD$\alpha$-classifier is observed when discriminating the Cauchy location alternative (Figure~\ref{simulationsFigure34}, left), but it is still close to the DD-classifiers. This can be attributed to the lower robustness of the zonoid depth.
However, when scaling enters the game (Cauchy location-scale alternative, Figure~\ref{simulationsFigure34}, right), the DD$\alpha$-classifier again performs quite satisfactorily.
The same picture arises when considering contaminated normal settings (Figure~\ref{simulationsFigure56}, left and right).
Under the location alternative, the DD$\alpha$-classifier is a bit worse than the DD-classifiers,  while it slightly outperforms them in a location-scale setting.

The relative robustness of the DD$\alpha$-classifier may be explained by two of its features: First it
maps the original data points to a compact set, the $q$-dimensional unit hypercube. Second, for classification in the unit hypercube, it employs the $\alpha$-procedure, which, by choosing a median angle in each step, is rather insensitive to outliers.

 Under exponential alternatives (Figure~\ref{simulationsFigure78}, left and right) the DD$\alpha$-classifier shows excellent performance, which is even similar to that of the the $k$-NN
for both location and location-scale alternatives.
Its results for the asymmetric location alternative (Figure~\ref{simulationsFigure910}, left) are somewhat ambiguous, though still close to those of the  DD-classifiers.
Concerning the normal-exponential alternative (Figure~\ref{simulationsFigure910}, right) the DD$\alpha$-classifier performs distinctly better than the others considered here.

On the basis of the simulation study we conclude: The DD$\alpha$-classifier (1) performs quite well under various settings of elliptically distributed alternatives, it (2) is rather robust to outlier prone data, and (3) shows a distinctly good behavior under the asymmetrically distributed alternatives considered and when the two classes originate from different families of distributions.

\subsection{Speed of the DD$\alpha$-procedure}\label{compsSpeed}

To estimate the speed of the DD$\alpha$-classification we have quantified the total time of training and classification times under two simulation settings, a  shift and a location-shift alternative concerning $d$-variate normals (see Table~\ref{timesTable}, header), with various values of dimension $d$ and of total size of training classes $n$. An experiment consists of a training phase based on two samples (each of size $n/2$) and an evaluation phase, where 2500 points (1250 from each distribution) are classified.
Each experiment is performed 100 times, then the average computation time is determined.
All these computations have been conducted on
a single kernel of the processor Core i7-2600 (3.4 GHz) having enough physical memory.

Table~\ref{timesTable} exhibits the average computation times (in seconds, with the standard deviations in parentheses) under the two distributional settings and for different $d$ and $n$.
As it is seen from the table, the DD$\alpha$-classifier is very fast, in the learning phase as well as in classifying high amounts of data.
However, computation times increase considerably with the number of training points, which is due to the many calculations of zonoid depth needed.
With dimension $d$ computation time grows slower, which
may be explained as follows.
With increasing dimension of the data space, more points come to lie on the convex hull (thus having depth $=2/n$) or outside it
(thus having depth $=0$). The algorithm from \cite{DyckerhoffKM96} computes the depth of such points much faster than that of points having larger depths.

\begin{table}
\caption{Computing times of DD$\alpha$-classification, in seconds.}
\label{timesTable}
\begin{tabular}{|c|c|c|c|c|}
\hline
 & \multicolumn{4}{|l|}{N$(\text{{\bf 0}}_d, \text{{\bf I}}_d)$} \\
 & \multicolumn{4}{|l|}{N$(0.25\cdot\text{{\bf 1}}_d, \text{{\bf I}}_d)$} \\
\hline\hline
                    & $d=5$     & $d=10$    & $d=15$    & $d=20$ \\
\hline
 $n=200$            & 0.14      & 1.55      & 1.89      & 2.24  \\
                    & (0.00014) & (0.00014) & (-)       & (-)   \\
\hline
 $n=500$            & 1.04      & 10.37     & 12.58     & 14.14 \\
                    & (0.00046) & (0.00052) & (0.00062) & (-)   \\
\hline
 $n=1000$           & 5.33      & 42.54     & 53.66     & 59.18 \\
                    & (0.0012)  & (0.0014)  & (0.0017)  & (-)   \\
\hline\hline
 & \multicolumn{4}{|l|}{N$(\text{{\bf 0}}_d, \text{{\bf I}}_d)$} \\
 & \multicolumn{4}{|l|}{N$((0.25 \text{{\bf \space 0}}^{\prime}_{d-1})^{\prime}, 5\cdot\text{{\bf I}}_d)$} \\
\hline\hline
                    & $d=5$     & $d=10$    & $d=15$    & $d=20$ \\
\hline
 $n=200$            & 0.15      & 1.62      & 1.94      & 2.2       \\
                    & (0.00014) & (0.00016) & (0.00021) & (0.00027) \\
\hline
 $n=500$            & 1.09      & 11.33     & 14.44     & 15.18     \\
                    & (0.00044) & (0.00059) & (0.00079) & (0.0010)  \\
\hline
 $n=1000$           & 5.24      & 47.63     & 67.22     & 74.15     \\
                    & (0.0011)  & (0.0016)  & (0.0022)  & (0.0026)  \\
\hline
\end{tabular}
\end{table}

\section{Benchmark studies}\label{secbenchmark}

Concerning real data, we take benchmark examples from \cite{LiCAL11,DuttaG11a,DuttaG11b} to compare the performance of the DD$\alpha$-classifier  with respect to AMR (Section \ref {benchmarkNonpar}). In addition we use four real data sets
from the UCI machine learning repository \cite{AsuncionN07}
to contrast the DD$\alpha$-classifier with the support vector machine (SVM) of  \cite{Vapnik98} regarding both performance and time (Section \ref {benchmarkSVM}).

\subsection{Benchmark comparisons with nonparametric classifiers}\label{benchmarkNonpar}

As our benchmark examples are well known, we refer to the literature for their detailed description and restrict ourselves to mentioning the dimension $d$, the number of classes $q$, the number of points used for training (\# train), the number of testing points (\# test) and the total number of points (\# total); see Table~\ref{datasetsTable}.

\begin{table}
\caption{Overview of benchmark examples; dimension ($d$), number of classes ($q$), number of training points (\# train), number of testing points (\# test), total number of points (\# total).}
\label{datasetsTable}
\begin{tabular}{|l|c|c|c|c|c|c|c|}
\hline
No.  & Dataset                                   & Results                                              & $q$ & $d$ & \# train & \# test & \# total \\ \hline\hline
1    & Biomedical                                & Tables~\ref{benchmarkTable1},~\ref{benchmarkTable3}  & 2 & 4  & 150 & 44   & 194  \\ \cline{3-8}
     &                                           & Table~\ref{benchmarkTable2}                          & 2 & 4  & 100 & 94   & 194  \\ \hline
2    & Blood                                     & Table~\ref{benchmarkTable2}                          & 2 & 3  & 374 & 374  & 748  \\ \cline{3-8}
     & Transfusion                               & Table~\ref{benchmarkTable3}                          & 2 & 3  & 500 & 248  & 748  \\ \hline
3    & Diabetes (1)                              & Table~\ref{benchmarkTable2}                          & 3 & 5  & 100 & 45   & 145  \\ \hline
4    & Diabetes (2)                              & Table~\ref{benchmarkSVMTable}                        & 2 & 8  & 767 & 1    & 768  \\ \hline
5    & Ecoli                                     & Table~\ref{benchmarkSVMTable}                        & 3 & 7  & 271 & 1    & 272  \\ \hline
6    & Glass                                     & Tables~\ref{benchmarkTable1},~\ref{benchmarkTable2}  & 2 & 5  & 100 & 46   & 146  \\ \cline{3-8}
     &                                           & Table~\ref{benchmarkSVMTable}                        & 2 & 9  & 145 & 1    & 146  \\ \hline
7    & Hemophilia                                & Table~\ref{benchmarkTable2}                          & 2 & 2  & 50  & 25   & 75   \\ \hline
8    & Image Segmentation                        & Table~\ref{benchmarkTable3}                          & 2 & 10 & 500 & 160  & 660  \\ \hline
9    & Iris                                      & Table~\ref{benchmarkSVMTable}                        & 3 & 4  & 149 & 1    & 150  \\ \hline
10   & Synthetic                                 & Tables~\ref{benchmarkTable1},~\ref{benchmarkTable2}  & 2 & 2  & 250 & 1000 & 1250 \\ \hline
\end{tabular}
\end{table}

Tables~\ref{benchmarkTable3},~\ref{benchmarkTable1} and ~\ref{benchmarkTable2} exhibit the performance
(in terms of AMR, with standard errors in parentheses) of the DD$\alpha$-classifier together with the performance of the different classifiers investigated in
\cite{LiCAL11}, \cite{DuttaG11a} and \cite{DuttaG11b} and based on the respective benchmark data.
When applying the DD$\alpha$-classifier an auxiliary procedure has to be chosen by which outsiders are treated.
In our benchmark study we employ several such procedures.

\begin{table}
\caption{Benchmark performance with DD- and other classifiers.}
\label{benchmarkTable3}
{\small
\begin{tabular}{|c|c|c|c|c|c|c|c|c|}
\hline
Dataset & LDA & QDA & $k$-NN & MM & MH & DM & DH & DD$\alpha$\\
\hline\hline
Biomedical   & 17.05 & 13.05 & 14.32 & 27.14 & 18.00 & 12.25 & 17.48 & 24.59\\
             & (0.49)& (0.38)& (0.45)& (0.6) & (0.49)& (0.4) & (0.51)& (0.63)\\
\hline
Blood        & 29.49 & 29.11 & 29.74 & 32.56 & 30.47 & 26.82 & 28.26 & 32.27\\
Transfusion  & (0.08)& (0.13)& (0.13)& (0.29)& (0.3) & (0.19)& (0.19)& (0.25)\\
\hline
Image        &  8.17 &  9.44 &  5.59 &  9.12 & 11.87 &  9.54 & 13.98 & 43.58\\
Segmentation & (0.2) & (0.19)& (0.19)& (0.23)& (0.25)& (0.2) & (0.29)& (0.34)\\
\hline
\end{tabular}
}
\end{table}

In  Table~\ref{benchmarkTable3} the DD$\alpha$-procedure is contrasted with the real data results in  \cite{LiCAL11}. Here
we use the same settings as in Section~\ref{compsPerfm} and classify the outsiders on a random basis.
All results in Table~\ref{benchmarkTable3} have been recalculated.

As we see from the Table, the performance of our new classifier is mostly worse than the classifiers considered in \cite{LiCAL11}. Only in the Blood Transfusion case the AMR has comparable size.
However, in this comparison
the eventual presence and treatment of outsiders plays a decisive role.
Observe that \cite{LiCAL11} in their procedures MH and DH use the random Tukey depth \cite{CuestaANR08} to
approximate the halfspace depth of a data point in dimension three and more.
But the random Tukey depth generally overestimates the halfspace depth so that some of the outsiders remain undetected.
This implies that, in the procedures MH and DH, considerably fewer points (we observed around 16\%, 4\% and 11\% correspondingly)
are treated as outsiders and assigned on a random basis.

In fact, as exactly determined by calculating the zonoid depth, the rate of outsiders in the Biomedical Data (with $d=4$) totals some 35\%,  in the Blood Transfusion Data ($d=3$) about 11\%,  and in the Image Segmentation Data with $d=10$ about 86\%.
This is in line with our expectation: the higher the dimension of the data the higher is the outsider rate.
In contrast to the MH and DH procedures, the DD$\alpha$-procedure detects all outsiders and, in the comparison of Table~\ref{benchmarkTable3}, assigns them randomly. Obviously the performance of the latter can be improved with a proper non-random procedure of outsider assignment.
In the subsequent benchmark comparisons several such procedures of non-random outsider assignment are included.

Dutta and Ghosh \cite{DuttaG11a} introduce classification based on projection depth and compare it with several variants of
the maximum-Mahalanobis-depth (MD)classifier.
The same authors \cite{DuttaG11b} propose an $L_p$-depth classifier (with optimized $p$) and contrast it with two types of MD.
To compare the DD$\alpha$-classifier on a par with \cite{DuttaG11a,DuttaG11b} we implement the following rules for handling outsiders:
First, $k$-nearest-neighbor rules are used with various $k$ and either Euclidean or Mahalanobis distance, the latter with moment or, alternatively, MCD estimates. Second, maximum Mahalanobis depth is employed, again based on moment or MCD estimation. As the $k$-NN results of the benchmark examples do not vary much with $k$, we restrict to $k=1$. (However, the performance of the classifiers can be improved by an additional cross-validation over $k$.) Consequently, five different rules for treating outsiders remain for comparison.
Tables~\ref{benchmarkTable1} and~\ref{benchmarkTable2} exhibit the performance of the DD$\alpha$-classifier \textit{vs.} the projection-depth classifiers of \cite{DuttaG11a} and the
$L_p$-depth classifiers of \cite{DuttaG11b}, respectively, regarding the benchmark examples investigated in these papers.
The last five columns of Table ~\ref{benchmarkTable1} and the bottom part of Table ~\ref{benchmarkTable2} report the AMR (standard deviations in parentheses) of the DD$\alpha$-classifier when one of the five outsiders treatments is chosen. The remaining columns are adopted as they stand in \cite{DuttaG11a} and \cite{DuttaG11b}.

\begin{table}
\caption{Benchmark comparison with projection depth classifiers.}
\label{benchmarkTable1}
\begin{tabular}{|c|c|c|c|c|c|c|}
\hline
Dataset & MD & MD & MD$_{\frac{3}{4}}$ & MD$_{\frac{3}{4}}$ & PD & PD \\
 & (SS) & (MS) & (SS) & (MS) & (SS) & (MS) \\
\hline\hline
Synthetic  & 13.00 & 11.60 & 10.30 & 10.40 & 10.00 & 10.50  \\
\hline
Glass      & 26.59 & 26.14 & 24.92 & 24.43 & 25.70 & 25.24  \\
           & (0.25)& (0.25)& (0.25)& (0.25)& (0.34)& (0.33) \\
\hline
Biomedical & 12.44 & 12.04 & 14.25 & 14.03 & 12.37 & 12.18  \\
           & (0.13)& (0.12)& (0.13)& (0.14)& (0.14)& (0.13) \\
\hline
\end{tabular}
\begin{tabular}{|c|c|c|c|c|c|}
\hline
Dataset & \multicolumn{5}{|c|}{\bf DD$\alpha$-classifier}\\
\cline{2-6}
& \multicolumn{3}{|c|}{1-NN} & \multicolumn{2}{|c|}{Mahalanobis}\\
\cline{2-4}
& Eucl. & \multicolumn{2}{|c|}{Mah. dist.} & \multicolumn{2}{|c|}{depth}\\
\cline{3-6}
 & dist. & Mom. & MCD & Mom. & MCD\\
\hline\hline
Synthetic  & 12.10 & 11.90 & 12.00 & 11.90 & 12.00\\
\hline
Glass      & 29.45 & 25.79 & 24.73 & 30.09 & 35.06\\
           & (0.20)& (0.17)& (0.18)& (0.18)& (0.22)\\
\hline
Biomedical & 13.51 & 19.59 & 17.90 & 12.91 & 15.23\\
           & (0.14)& (0.18)& (0.17)& (0.14)& (0.16)\\
\hline
\end{tabular}
\end{table}

Regarding the Biomedical Data, \cite{DuttaG11a} do not specify the sample sizes they use in training and testing.
For the DD$\alpha$-classifier, we select 100 observations of the larger class and 50 of the smaller class to form the training sample; the remaining observations constitute the testing sample.
As it is seen from Table~\ref{benchmarkTable1} the DD$\alpha$-classifier shows results similar to the projection-depth classifier (except with the Synthetic Data), while the performance of outsider-handling methods varies depending on the type of the data.
Specifically, with the Glass Data 1-NN based on the Mahalanobis distance (both with the moment and the robust estimate) performs best in handling outsiders.
On the other hand, with the Biomedical Data the same approach performs quite poorly, while treating outsiders with
moment-estimated Mahalanobis depth or Euclidean 1-NN yields best results.

\begin{table}
\caption{Benchmark comparison with L$_p$-depth classifiers.}
\label{benchmarkTable2}
{\small
\begin{tabular}{|c|c|c|c|c|c|c|c|c|c|}
\hline
Data- & \multicolumn{2}{|c|}{} & \multicolumn{2}{|c|}{} & \multicolumn{5}{|c|}{\bf DD$\alpha$-classifier}\\
\cline{6-10}
set & \multicolumn{2}{|c|}{} & \multicolumn{2}{|c|}{} & \multicolumn{3}{|c|}{1-NN} & \multicolumn{2}{|c|}{Mahalanobis}\\
\cline{6-8}
& \multicolumn{2}{|c|}{MD} & \multicolumn{2}{|c|}{L$_p$D} & Eucl. & \multicolumn{2}{|c|}{Mah. dist.} & \multicolumn{2}{|c|}{depth}\\
\cline{2-5}\cline{7-10}
& Mom. & MCD & Mom. & MCD & dist. & Mom. & MCD & Mom. & MCD\\
\hline\hline
Syn.        & 10.20 & 10.60 &  9.60 & 10.70 & 12.10 & 11.90 & 12.00 & 11.90 & 12.00\\
            & & & & & & & & & \\
\hline
Hem.        & 15.84 & 17.13 & 15.39 & 16.43 & 16.63 & 17.98 & 18.36 & 18.65 & 19.39\\
            & (0.30)& (0.32)& (0.32)& (0.32)& (0.20)& (0.20)& (0.19)& (0.22)& (0.22)\\
\hline
Gla.        & 26.80 & 24.80 & 27.64 & 24.75 & 30.13 & 28.37 & 26.63 & 32.88 & 36.82\\
            & (0.26)& (0.29)& (0.29)& (0.26)& (0.19)& (0.22)& (0.20)& (0.22)& (0.23)\\
\hline
Biom.        & 12.35 & 14.48 & 12.68 & 15.11 & 13.74 & 22.09 & 20.89 & 14.34 & 17.28\\
            & (0.14)& (0.15)& (0.15)& (0.15)& (0.09)& (0.16)& (0.14)& (0.12)& (0.14)\\
\hline
Diab.       &  8.22 & 11.49 &  9.39 & 11.92 & 10.77 & 18.36 & 18.33 & 12.70 & 15.90\\
            & (0.18)& (0.22)& (0.21)& (0.27)& (0.12)& (0.18)& (0.20)& (0.18)& (0.19)\\
\hline
B.Tr.       & 22.75 & 22.17 & 22.30 & 22.06 & 23.11 & 22.73 & 22.92 & 22.59 & 22.17\\
            & (0.07)& (0.08)& (0.07)& (0.07)& (0.06)& (0.06)& (0.06)& (0.06)& (0.06)\\
\hline
\end{tabular}
}
\end{table}

Table~\ref{benchmarkTable2} presents a similar comparison of the DD$\alpha$-classifier with the $L_p$-classifier of \cite{DuttaG11b}.
The same approaches are included to treat outsiders.
In all six benchmark examples the DD$\alpha$-classifier generally performs worse than the best $L_p$-depth classifier.
However, its performance substantially depends on the chosen treatment of outsiders.
In all examples
the AMR of the DD$\alpha$-classifier comes close to that of the $L_p$-depth classifier, provided the outsider treatment is properly selected.
On the Hemophilia Data, e.g., Euclidean 1-NN should be chosen.
On the Glass Data a 1-NN outsider treatment with robust Mahalanobis distance performs relatively best, etc.
On the Blood Transfusion Data all outsider-handling approaches show equally good performance, which appears to be typical when $n$ is relatively large
compared to $d$.

\subsection{Benchmark comparisons with SVM}\label{benchmarkSVM}

The support vector machine (SVM) is a powerful solver of the classification problem and has been widely used in applications.
However, different from the DD$\alpha$-classifier, the SVM is a parametric approach, as in applying it certain parameters have to be adjusted: the box-constraint and the kernel parameters. The AMR performance of the SVM depends heavily on the choice of these parameters. In applications, optimal parameters are selected by some cross-validation, which affords extensive calculations.
Once these parameter have been optimized, SVM-classification is usually very fast and precise.

In comparing the SVM with the DD$\alpha$-procedure, this step of parameter optimization has to be somehow accounted for. Here we introduce a two-fold view on the comparison problem: Two values of the AMR are calculated, first the \textit{best AMR} when the parameters have been optimally selected, second the \textit{expected AMR} when the parameters are systematically varied over specified ranges. Corresponding training times are also clocked. As ranges we choose the intervals between the smallest and the largest number that arise as an optimal value in one of our benchmark data examples. This seems us a fair and, regarding the parameter ranges, rather conservative approach.

As benchmark four well-known data sets are employed in the sequel, Diabetes, Ecoli, Glass, and Iris Data being taken  \cite{AsuncionN07}. Following \cite{DuttaG11a} the two biggest classes of the Glass Data have been selected, and similarly to \cite{DuttaG11b} we have chosen three of the bigger classes from the Ecoli Data. The DD$\alpha$-classifier is calculated with the same outsider treatments as above. For the SVM-classifier we use radial basis function kernels as implemented in LIBSVM with the R-Package ``e1071'' as an R-interface. Leave-one-out cross validation is employed for performance estimation of the all classifiers. The computation has been done on the same PC as in Section \ref {compsSpeed}.

The results on the best AMR together with time quantities and portions of outsiders are collected in the Table~\ref{benchmarkSVMTable}. The Iris Data appears twice in the Table. First the original are used, and second the same data after a preprocessing step. The preprocessing consists in the exclusion of an obvious outlier in the DD-plot that was identified by visual inspection of the plot.

\begin{table}
\caption{Benchmark comparison with the support vector machine; $\gamma$ - kernel parameter, $C$ - box constraint.}
\label{benchmarkSVMTable}
{\small
\begin{tabular}{|c|l|c|c|c|c|c|c|}
\hline
Data- & & \multicolumn{5}{|c|}{\bf DD$\alpha$-classifier} & {\bf SVM}\\
\cline{3-8}
set & & \multicolumn{3}{|c|}{1-NN} & \multicolumn{2}{|c|}{Mahalanobis} & \\
\cline{3-7}
& & Eucl. & \multicolumn{2}{|c|}{Mah. dist.} & \multicolumn{2}{|c|}{depth} & \\
\cline{4-7}
             & Legend       & dist.  & Mom.   & MCD    & Mom.    & MCD     & Opt. (CV)\\
\hline\hline
Diab.        & Error        & 28.26  & 30.6   & 34.51  & 24.35   & 31.77   & 23.18           \\
             & Time:train   & 16.63  & 16.62  & 16.59  & 16.58   & 17.39   & 0.05 (875)      \\
             & Time:test    & 0.033  & 0.009  & 0.0092 & 0.0035  & 0.0037  & 0.0023          \\
             & $\gamma/C$   &        &        &        &         &         & 0.056/1         \\
             & \% outsiders & 62.24  & 62.24  & 62.24  & 62.24   & 62.24   & \\
\hline
Ecoli        & Error        & 10.29  & 11.4   & 12.13  & 12.13   & 16.18   & 3.68            \\
             & Time:train   & 0.26   & 0.26   & 0.26   & 0.26    & 0.26    & 0.0077 (105)    \\
             & Time:test    & 0.014  & 0.0026 & 0.0032 & 0.001   & 0.00044 & 0.0019          \\
             & $\gamma/C$   &        &        &        &         &         & 5.62/1.78       \\
             & \% outsiders & 75     & 75     & 75     & 75      & 75      & \\
\hline
Glass        & Error        & 18.49  & 26.03  & 31.51  & 34.93   & 34.93   & 21.23           \\
             & Time:train   & 0.31   & 0.32   & 0.31   & 0.32    & 0.32    & 0.0082 (36)     \\
             & Time:test    & 0.0083 & 0.0019 & 0.0016 & 0.00014 & 0.00055 & 0.0024          \\
             & $\gamma/C$   &        &        &        &         &         & 0.56/1          \\
             & \% outsiders & 95.89  & 95.89  & 95.89  & 95.89   & 95.89   & \\
\hline
Iris         & Error        & 37.33  & 37.33  & 37.33  & 36      & 46.67   & 4.67            \\
             & Time:train   & 0.07   & 0.07   & 0.07   & 0.07    & 0.07    & 0.0051 (30)     \\
             & Time:test    & 0.0046 & 0.0018 & 0.0013 & 0.00033 & 0.00047 & 0.0017          \\
             & $\gamma/C$   &        &        &        &         &         & 0.056/10        \\
             & \% outsiders & 50     & 50     & 50     & 50      & 50      & \\
\hline
Iris         & Error        & 3.36   & 3.36   & 4.03   & 2.68    & 13.42   & 2.68            \\
(Pre.)       & Time:train   & 0.07   & 0.07   & 0.07   & 0.07    & 0.07    & 0.0052 (30)     \\
             & Time:test    & 0.0046 & 0.0011 & 0.0013 & 0.0006  & 0.00027 & 0.0017          \\
             & $\gamma/C$   &        &        &        &         &         & 0.1/3.16        \\
             & \% outsiders & 51.68  & 51.68  & 51.68  & 51.68   & 51.68   & \\
\hline
\end{tabular}
}
\end{table}

The overall analysis of the Table~\ref{benchmarkSVMTable} shows that, even if using an arbitrary technique for handling outsiders, the DD$\alpha$-classifier mostly performs not much worse than an SVM where the parameters have been optimally chosen. In contrast, if the SVM is employed with some non-optimized parameters, its AMR can be considerably larger than that of the DD$\alpha$-classifier. For the regarded data sets average errors of the SVM over the relevant intervals varied from 44.99\% to 66.67\% (not reported in the Table).

The times needed to classify a new object (also given in Table~\ref{benchmarkSVMTable}) are quite comparable. But as the parameters of the SVM have to be adjusted first by running it many times for cross-validation, the computational burden of its training phase is much higher than that of the DD$\alpha$-classifier, which has to be run only once. Recall that the latter is nonparametric regarding tuning parameters. For example, in our implementation it took 875 seconds to determine approximate optimal values of SVM parameters for the Diabetes Data and similarly substantial times for the others (see Table~\ref{benchmarkSVMTable}, in parentheses).


\section{Discussion and conclusions} \label{concl}

A new classification procedure has been proposed that is completely nonparametric.
The DD$\alpha$-classifier transforms the $d$-variate data to a $q$-variate depth plot and performs linear classification in an extended depth space.
The depth transformation is done by the zonoid depth, and the final classification by the $\alpha$-procedure.
The procedure has attractive properties: First, it proves to be very fast and efficient in the training as well as in the testing phase; in this it highly outperforms existing alternative nonparametric classifiers, and also - regarding the training phase - the support vector machine.
Second, in many settings of elliptically distributed alternatives, its AMR is of similar size than that of the competing classifiers.
Moreover, it is rather robust to outlier prone data.
As a nonparametric approach, the new procedure shows a particularly good behavior under asymmetrically distributed alternatives and, in certain cases, when the two classes originate from different families of distributions.
Other than many competitors, it considers all classes in the multi-class classification problem even when performing binary classification.
Different for KNN, SVM and other kernel based procedures our method does not need to be parametrically tuned.
Also several theoretical properties of the DD$\alpha$-procedure have been derived:
It operates in a rather simple way if the data generating processes are elliptical, and a Bayes rule holds if $q=2$ and the two classes are mirror symmetric.

The zonoid depth has many theoretical and computational advantages: Most important here, it is efficiently computed also in higher dimensions. However, as it takes its maximum at the mean of the data, the zonoid depth lacks robustness.
Nevertheless, the DD$\alpha$-classifier shows a rather robust behavior. Its relative robustness can be explained as follows: The original data points are mapped to a compact set, the $q$-dimensional unit hypercube, and then classified by the $\alpha$-procedure. The latter, by choosing a median angle in each step, is rather insensitive to outliers.

Points that are not within the convex hull of at least one training set must be specially treated as their depth representation is zero.
To classify those so called outsiders several approaches have been used and compared. Instead of assigning them randomly, which disadvantages
the DD$\alpha$-procedure like other procedures based on halfspace or simplicial depth, one should classify outsiders by $1$-NN and some distance or by a properly chosen maximum depth rule.

To contrast the DD$\alpha$-procedure with an SVM approach, a novel way of comparison has been taken: An optimal performance of an SVM has been evaluated, that arises under an optimal choice of the parameters, as well as an average performance, where the parameters vary over specified conservative intervals.
It came out that, even with an arbitrary handling of outsiders, the  DD$\alpha$-classifier mostly performs not much worse than an SVM whose parameters have been optimally chosen. However, if the SVM is employed with some non-optimized parameters, the AMR can be considerably larger than that of the DD$\alpha$-classifier.

More investigations are needed on the consistency of the DD$\alpha$-classifier, its behavior on skewed or fat-tailed data, the - possibly adaptive - choice of outsider treatments, and the use of alternative notions of data depth. These are intended for future research.

\begin{acknowledgements}
Thanks are to Rainer Dyckerhoff for his constructive remarks on the paper as well as to the other participants of the Witten Workshop on
``Robust methods for dependent data'' for discussions. The helpful suggestions of two referees are gratefully acknowledged.
\end{acknowledgements}



\end{document}